\documentclass[10pt,twocolumn,letterpaper]{article}

\usepackage{cvpr}
\usepackage{times}
\usepackage{epsfig}
\usepackage{graphicx}
\usepackage{amsmath}
\usepackage{amssymb}
\usepackage{xcolor}
\usepackage{siunitx}
\usepackage{colortbl}
\usepackage{authblk}
\usepackage[T1]{fontenc}
\usepackage{mathabx}
\usepackage[labelfont=small, textfont=small]{caption}
\usepackage[skip=0cm,list=true]{subcaption}
\usepackage{multirow}
\usepackage{enumitem}

\usepackage[pagebackref=true,breaklinks=true,letterpaper=true,colorlinks,bookmarks=false]{hyperref}
\cvprfinalcopy
\def\cvprPaperID{5}

\ifcvprfinal\fi
\begin{document}

\title{ViPR: Visual-Odometry-aided Pose Regression for 6DoF Camera Localization}

\author[1]{Felix Ott}
\author[1,2]{Tobias Feigl}
\author[1,2]{Christoffer Löffler}
\author[1,3]{Christopher Mutschler}
\affil[1]{Fraunhofer Institute for Integrated Circuits IIS, Nuremberg, Germany}
\affil[2]{Department of Computer Science, FAU Erlangen-Nuremberg, Germany}
\affil[3]{Department of Statistics, Ludwig-Maximilians-University (LMU), Munich, Germany}
\affil[ ]{\tt\small {\{felix.ott | tobias.feigl | christoffer.loeffler | christopher.mutschler\}@iis.fraunhofer.de}}

\maketitle

\thispagestyle{empty}
\begin{abstract}%
Visual Odometry (VO) accumulates a positional drift in long-term robot navigation tasks. Although Convolutional Neural Networks (CNNs) improve VO in various aspects, VO still suffers from moving obstacles, discontinuous observation of features, and poor textures or visual information. While recent approaches estimate a 6DoF pose either directly from (a series of) images or by merging depth maps with optical flow (OF), research that combines absolute pose regression with OF is limited. 

We propose ViPR, a novel modular architecture for long-term 6DoF VO that leverages temporal information and synergies between absolute pose estimates (from PoseNet-like modules) and relative pose estimates (from FlowNet-based modules) by combining both through recurrent layers. Experiments on known datasets and on our own Industry dataset show that our modular design outperforms state of the art in long-term navigation tasks.
\end{abstract}%

\section{Introduction}\label{chapter_introduction}%

Real-time tracking of mobile objects (e.g., forklifts in industrial areas) allows to monitor and optimize workflows and tracks goods for automated inventory management. Such environments typically include large warehouses or factory buildings, and localization solutions often use a combination of radio-, LiDAR- or radar-based systems, etc.%

However, these solutions often require infrastructure or they are costly in their operation. An alternative approach is a (mobile) optical pose estimation based on ego-motion. Such approaches are usually based on SLAM (Simultaneous Localization and Mapping), meet the requirements of exact real-time localization, and are also cost-efficient.%

Available pose estimation approaches are categorized into three groups: classical, hybrid, and deep learning (DL)-based methods. Classical methods often require an infrastructure that includes either synthetic (i.e., installed in the environment) or natural (e.g., walls and edges) markers. The accuracy of the pose estimation depends to a large extent on suitable invariance properties of the available features such that they can be reliably recognized. However, to reliably detect features, we have to invest a lot of expensive computing time~\cite{marker,marker2}. Additional sensors (e.g., inertial sensors, depth cameras, etc.) or additional context (e.g., 3D models of the environment, prerecorded landmark databases, etc.) may increase the accuracy but also increase system complexity and costs~\cite{loeffler}. Hybrid methods~\cite{shotton_scene,brachman,brachman_dsac,guzman,valentin} combine geometric and machine learning (ML) approaches. For instance, ML predicts the 3D position of each pixel in world coordinates, from which geometry-based methods infer the camera pose~\cite{duong}.%

\begin{figure}[b!]%
    \centering%
    \vspace{-5mm}%
    \includegraphics[width=1\linewidth]{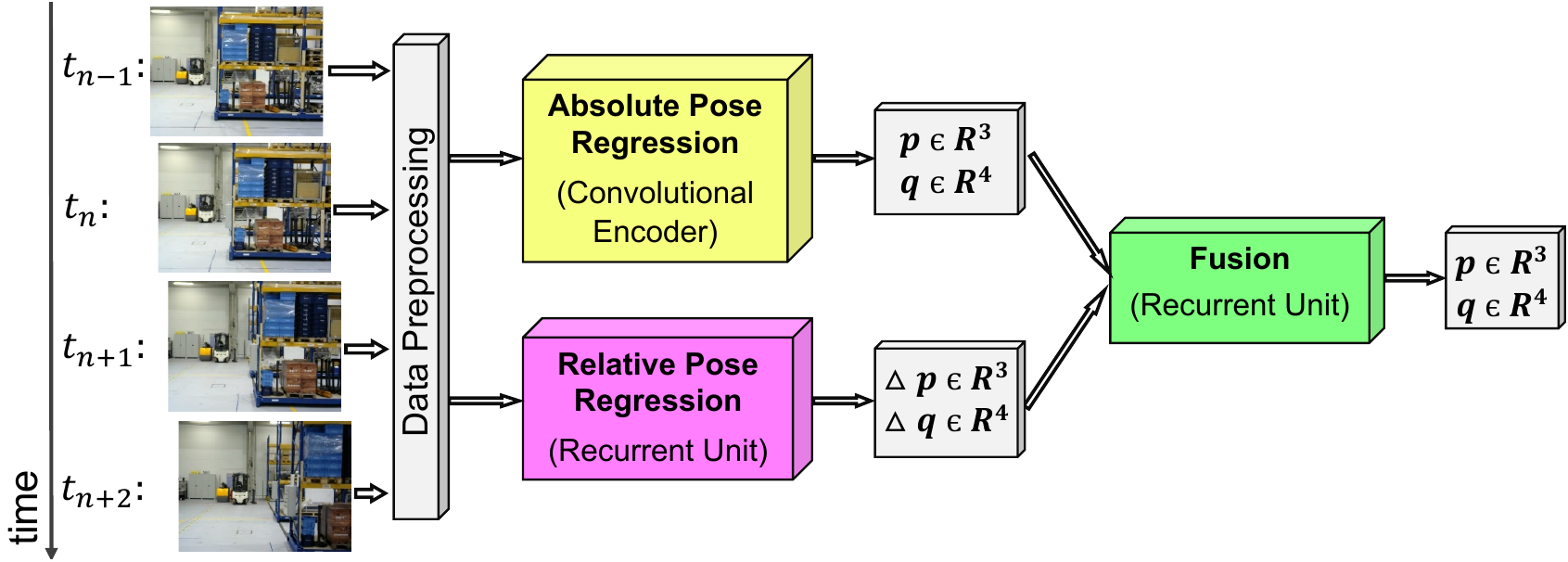}%
    \vspace{-2mm}%
    \caption{Our pose estimation pipeline solves the APR- and RPR-tasks in parallel, and recurrent layers estimate the final 6DoF pose.}%
    \label{image_overview_model}%
\end{figure}%

Recent DL approaches partly address the above mentioned issues of complexity and cost, and also aim for high positioning accuracy, e.g., regression forests~\cite{meng_regression_forests,valentin} learn a mapping of images to positions based on 3D models of the environment. Absolute pose regression (APR) uses DL~\cite{sattler_apr} as a cascade of convolution operators to learn poses only from 2D images. The pioneer \texttt{PoseNet}~\cite{kendall_pose} has been extended by Bayesian approaches~\cite{kendall_modelling}, long short-term memories (LSTMs)~\cite{walch_lstm} and others~\cite{hourglass,ctcnet,distancenet,gposenet}. Recent APR methods such as \texttt{VLocNet}~\cite{valada_deep,vlocnet2} and \texttt{DGRNets}~\cite{dgrnets} introduce relative pose regression (RPR) to address the APR-task. While APR needs to be trained for a particular scene, RPR may be trained for multiple scenes~\cite{sattler_apr}. However, RPR alone does not solve the navigation task.%

For applications such as indoor positioning, existing approaches are not yet mature, i.e., in terms of robustness and accuracy to handle real-world challenges such as changing environment geometries, lighting conditions, and camera (motion) artifacts. This paper proposes a modular fusion technique for 6DoF pose estimation based on a \texttt{PoseNet}-like module and predictions of a relative module for VO. Our novel relative module uses the flow of image pixels between successive images computed by \texttt{FlowNet2.0}~\cite{flownet2} to capture time dependencies in the camera movement in the recurrent layers, see Fig.~\ref{image_overview_model}. Our model reduces the positioning error using this multitasking approach, which learns both the absolute poses based on monocular (2D) imaging and the relative motion for the task of estimating VO.

We evaluate our approach first on the small-scale \texttt{7-Scenes}~\cite{shotton_scene} dataset. As other datasets are unsuitable to evaluate continuous navigation tasks we also release a dataset that can be used to evaluate various problems arising from real industrial scenarios such as inconsistent lighting, occlusion, dynamic environments, etc. We benchmark our approach on both datasets against existing approaches ~\cite{kendall_pose,walch_lstm} and show that we consistently outperform the accuracy of their pose estimates.%

The rest of the paper is structured as follows. Section~\ref{chapter_related_work} discusses related work. Section~\ref{chapter_model} provides details about our architecture. We discuss available datasets and introduce our novel \textit{Industry} dataset in Section~\ref{chapter_datasets}. We present experimental results in Section~\ref{chapter_results} before Section~\ref{chapter_conclusion} concludes.

\renewcommand{\slash}{/\penalty\exhyphenpenalty\hspace{0pt}}
\section{Related Work}\label{chapter_related_work}%

SLAM-driven 3D point registration methods enable precise self-localization even in unknown environments. Although VO has made remarkable progress over the last decade, it still suffers greatly from scaling errors of real and estimated maps~\cite{F6,F13,F14,F17,F18,F7,F8,F16,bergen_visual_2004,F15}. With more computing power, Visual Inertial SLAM combines VO with Inertial Measurement Unit (IMU) sensors to partly resolve the scale ambiguity, to provide motion cues without visual features~\cite{F6,F144,F17}, to process more features, and to make tracking more robust~\cite{F13,F18}. Multiple works combine global localization in a scene with SLAM/(Inertial) VO \cite{lynen, dutoit, mur-artal, schneider, geppert, middelberg, jones}. However, recent SLAM methods do not yet meet industry-strength with respect to accuracy and reliability~\cite{palmarini2017innovative, feigl_arcore} as they need undamaged, clean and undisguised markers~\cite{F15,F1003} and as they still suffer from long-term stability and the effects of movement, sudden acceleration and occlusion~\cite{F21}. SIFT-like point-based features~\cite{lowe} for the localization from landmarks~\cite{bergamo,hao_3d,li_worldwide,wang_sift} require efficient retrieval methods, use VLAD encodings such as \texttt{DenseVLAD}~\cite{torii}, use anchor points such as \texttt{AnchorNet}~\cite{anchornet}, or use RANSAC-based optimization such as \texttt{DSAC}~\cite{brachman_dsac} and \texttt{ActiveSearch}~\cite{sattler_activesearch}.

VO primarily addresses the problem of separating ego- from feature-motion and suffers from area constraints, poorly textured environments, scale drift, a lack of an initial position, and thus inconsistent camera trajectories~\cite{cadena2016past}. Instead, \texttt{PoseNet}-like architectures (see Sec.~\ref{subsection:RW_APR}) that estimate absolute poses on single-shot images are more robust, less compute-intensive, and can be trained in advance on application data. Unlike VO, they do not suffer from a lack of initial poses and do not require access to camera parameters, good initialization, and handcrafted features~\cite{seifi}. Although the joint estimation of relative poses may contribute to increasing accuracy (see Sec.~\ref{subsection:RW_APR-HPR_hybrids}), such hybrid approaches still suffer from dynamic environments, as they are often trained offline in quasi-rigid environments. While optical flow (see Sec.~\ref{subsection:RW_OF}) addresses these challenges it has not yet been combined with APR for 6DoF self-localization.%

\subsection{Absolute Pose Regression (APR)}\label{subsection:RW_APR}%

Methods that derive a 6DoF pose directly from images have been studied for decades. Therefore, there are currently many classic methods whose complex components are replaced by machine learning (ML) or DL. For instance, \texttt{RelocNet}~\cite{relocnet} learns metrics continuously from global image features through a camera frustum overlap loss. \texttt{CamNet}~\cite{camnet_ding} is a coarse (image-based)-to-fine (pose-based) retrieval-based model that includes relative pose regression to get close to the best database entry that contains extracted features of images. \texttt{NNet}~\cite{laskar} queries a database for similar images to predict the relative pose between images and a RANSAC~\cite{sweeney} solves the triangulation to provide a position. While those \textit{classic} approaches have already been extended with DL-components their pipelines are expensive as they embed feature matching and projection and/or manage a database. Most recent (and simple) DL-based also outperform their accuracies.%

The key idea of \texttt{PoseNet}~\cite{kendall_pose} and its variants~\cite{kendall_geometric,kendall_modelling,gal,walch_lstm,walch_MA,deepvo_vo,contextualnet,seifi,naseer,shotton_scene} among others such as \texttt{BranchNet}~\cite{naseer} and \texttt{Hourglass}~\cite{shotton_scene} is to use a CNN for camera (re-)localization. \texttt{PoseNet} works with scene elements of different scales and is partially insensitive to light changes, occlusions and motion blur. However, while \texttt{Dense~PoseNet}~\cite{kendall_pose} crops subimages, \texttt{PoseNet2}~\cite{kendall_geometric} jointly learns network and loss function parameters, \cite{kendall_modelling} links a \emph{Bernoulli} function and applies variational inference~\cite{gal} to improve the positioning accuracy. However, those variants work with single images, and hence, do not use the temporal context (which is available in continuous navigation tasks), that could help to increase accuracy.%

In addition to \texttt{PoseNet+LSTM}~\cite{walch_lstm}, there are also similar approaches that exploit time-context that is inherently given by consecutive images (i.e., \texttt{DeepVO}~\cite{deepvo_vo}, \texttt{ContextualNet}~\cite{contextualnet}, and \texttt{VidLoc}~\cite{clark_vidloc}). Here, the key-idea is to identify temporal connections in-between the feature vectors (extracted from images) with LSTM-units and to only track feature correlations that contribute the most to the pose estimation. However, there are hardly any long-term dependencies between successive images, and therefore LSTMs give worse or equal accuracy to, for example, simple averaging over successively estimated poses~\cite{seifi}. Instead, we combine estimated poses from time-distributed CNNs with estimates of the OF to maintain the required temporal context in the features of image series.%

\subsection{APR/RPR-Hybrids}
\label{subsection:RW_APR-HPR_hybrids}%

In addition to approaches that derive a 6DoF pose directly from an image there are hybrid methods that combine them with VO to increase the accuracy. \texttt{VLocNet}~\cite{valada_deep} is closely related to our approach as it estimates a global pose and combines it with VO (but it does not use OF). To further improve the (re-)localization accuracy \texttt{VLocNet++}~\cite{vlocnet2} uses features from a semantic segmentation. However, we use different networks and do not need to share weights between VO and the global pose estimation. \texttt{DGRNets}~\cite{dgrnets} estimates both the absolute and relative poses, concatenates them, and uses recurrent CNNs to extract temporal relations between consecutive images. This is similar to our approach but we estimate the relative motion with OF, which allows us to train in advance on large datasets, making the model more robust. \texttt{MapNet}~\cite{mapnet} learns a map representation from input data, combines it with GPS, inertial data, and unlabeled images, and uses pose graph optimization (PGO) to combine absolute and relative pose predictions. However, compared to all other methods the most accurate extension of it, \texttt{MapNet+PGO}, does not work on purely visual information, but exploits additional sensors.%

\begin{figure}[t!]
	\begin{minipage}[t]{0.325\linewidth}
		\centering
		\includegraphics[width=0.95\linewidth]{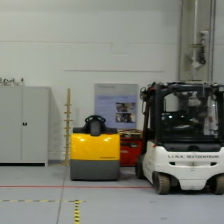}
	\end{minipage}
	\hfill
	\begin{minipage}[t]{0.325\linewidth}
		\centering
		\includegraphics[width=0.95\linewidth]{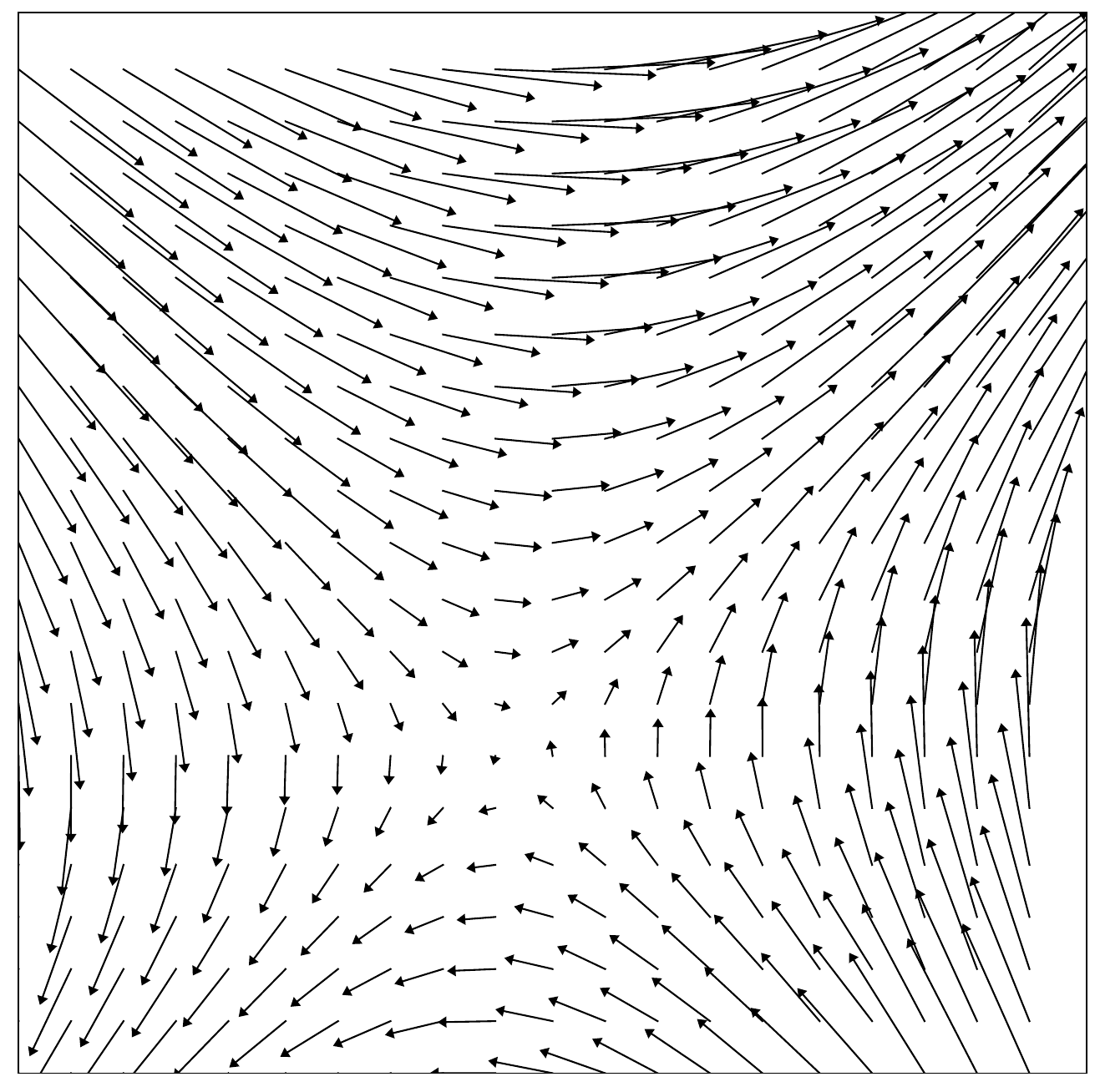}
	\end{minipage}
	\hfill
	\begin{minipage}[t]{0.325\linewidth}
		\centering
		\includegraphics[width=0.95\linewidth]{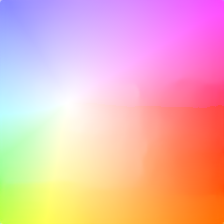}
	\end{minipage}
	\vspace{-7mm}
	\caption{Optical flow (OF): input image (left); OF-vectors as RPR-input (middle); color-coded visualization of OF~\cite{baker_dataset} (right).}
	\label{image_of_intro}
	\vspace{-0.5cm}
\end{figure}

\subsection{Optical Flow}%
\label{subsection:RW_OF}%

Typically, VO uses OF to extract features from image sequences. Motion fields, see Fig.~\ref{image_of_intro} (middle), are used to estimate trajectories of pixels in a series of images. For instance, \texttt{Flowdometry}~\cite{flowdometry} and \texttt{LS-VO}~\cite{constante} estimate displacements and rotations from OF. \cite{mansur} proposed a VO-based dead reckoning system that uses OF to match features. \cite{zhao} combined two CNNs to estimate the VO-motion: \texttt{FlowNet2-ss}~\cite{flownet2} estimates the OF and PCNN~\cite{pcnn} links two images to process global and local pose information. However, to the best of our knowledge, we are the first to propose an OF-based architecture that estimates the relative camera movement through RNNs, using OF \cite{flownet2}.

\section{Proposed Model}\label{chapter_model}%

After a data preprocessing that crops subimages of size $224 \times 224 \times 3$ from a sequence of four images, our pose regression pipeline consists of three parts (see Fig.~\ref{image_fusion2}): an APR-network, a RPR-network, and a 6DoF pose estimation (PE) network. PE uses the outputs of the APR- and RPR-networks to provide the final 6DoF pose.%

\subsection{Absolute Pose Regression (APR) Network}\label{chapter_model_absnet}%

Our APR-network predicts the 6DoF camera pose from three input images based on the original \texttt{PoseNet}~\cite{kendall_pose} model (i.e., essentially a modified GoogLeNet~\cite{googlenet} with a regression head instead of a softmax) to train and predict the absolute positions $\textbf{\textit{p}} \in \mathbb{R}^{3}$ in the Euclidean space and the absolute orientations $\textbf{\textit{q}} \in \mathbb{R}^{4}$ as quaternions. From a single monocular image \textit{I} the model predicts the pose

\begin{figure*}[t!]%
     \centering%
     \includegraphics[width=1\linewidth]{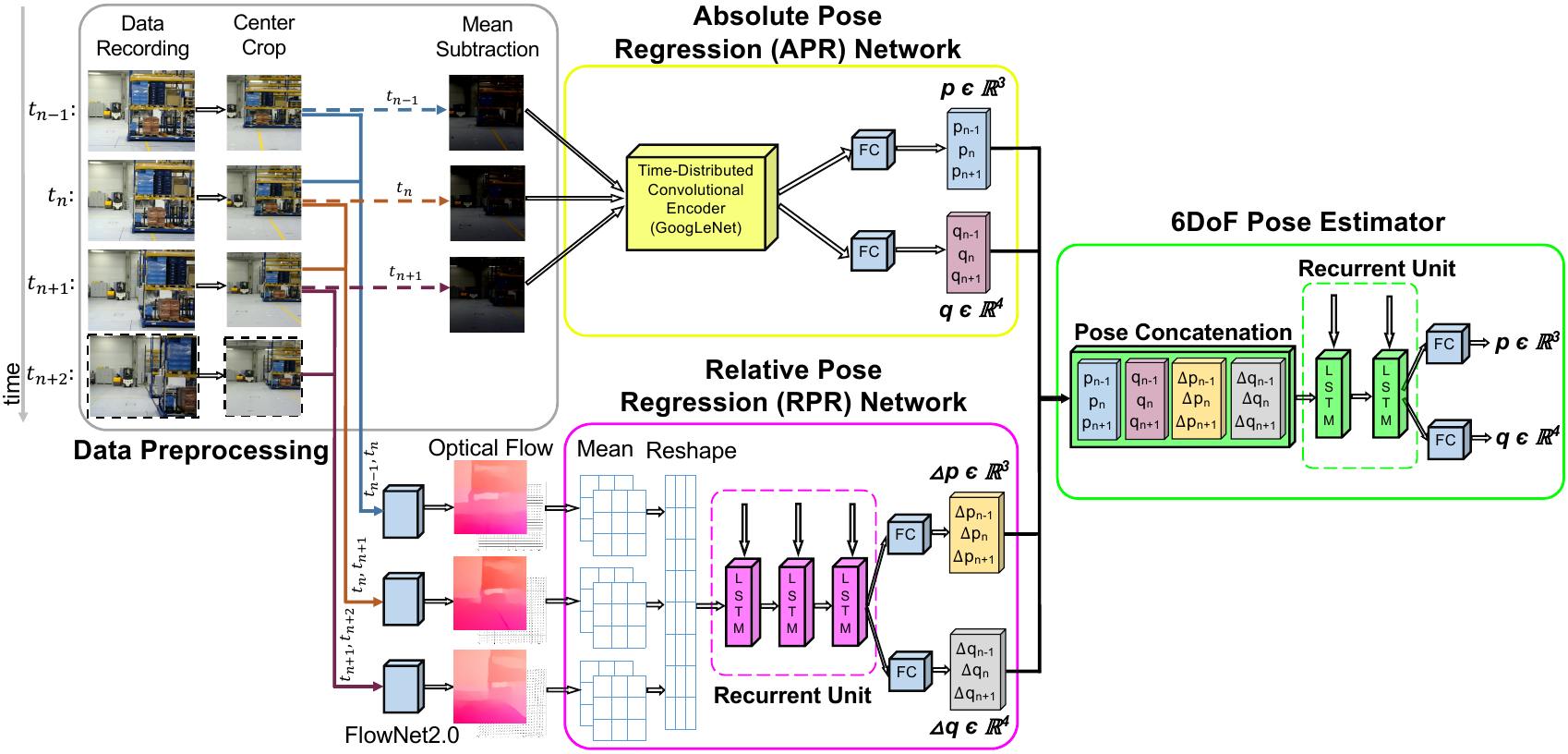}%
     \vspace{-2mm}%
     \caption{\textbf{Pipeline of the ViPR-architecture.} Data preprocessing (grey): Four consecutive input images ($t_{n-1}, \dots, t_{n+2}$) are center cropped. For the \textit{absolute} network the mean is subtracted. For the \textit{relative} network the OF is precomputed by FlowNet2.0~\cite{flownet2}. The \textit{absolute} poses are predicted by our \textit{time-distributed} APR-network (yellow). The RPR-network (purple) predicts the transformed \textit{relative} displacements and rotations on reshaped mean vectors of the OF with (stacked) LSTM-RNNs. The PE modules (green) concatenates the absolute and relative modules and predicts the absolute 6DoF poses with stacked LSTM-RNNs.}%
     \vspace{-4mm}%
    \label{image_fusion2}%
\end{figure*}

\vspace{-3mm}

\begin{equation}
	\tilde{\textbf{\emph{x}}} = [\tilde{\textbf{\emph{p}}}, \tilde{\textbf{\emph{q}}}],
	\label{equ_posenet_pose}
\end{equation}

as approximations to the actual $\textbf{\textit{p}}$ and $\textbf{\textit{q}}$. As the original model learns the image context, based on shape and appearance of the environment, but does not exploit the time context and relation between consecutive images~\cite{kendall_geometric}, we adapted the model to a \textit{time-distributed} variant. Hence, instead of a single image the new model receives three (consecutive) input images (at timesteps $t_{n-1}$, $t_n$, and $t_{n+1}$), see top part of Fig.~\ref{image_fusion2}, uses three separate dense layers (one for each pose) with 2,048 neurons each, and each of the dense layers yields a pose. The middle pose yields the most accurate position for the image at time step $t_n$.

\subsection{Relative Pose Regression (RPR) Network}\label{chapter_model_relnet}%

Our RPR-network uses \texttt{FlowNet2.0}~\cite{flownet2} on each consecutive pairs of the four input images to compute an approximation of the OF (see Fig.~\ref{image_of_intro}) and to predict three relative poses for later use. As displacements of similar length but from different camera viewing directions result in different OFs, the displacement and rotation of the camera between pairwise images must be \textit{relative} to the camera's viewing direction of the first image. Therefore, we transform each camera's global coordinate systems $(x_n,y_n,z_n)$ to the same local coordinate system $(\tilde{x}_n,\tilde{y}_n,\tilde{z}_n)$ by

\vspace{-2mm}

\begin{equation}\label{equ_backtransformation1}
	\begin{pmatrix}
	\tilde{x}_n \\
	\tilde{y}_n \\
	\tilde{z}_n \\
	\end{pmatrix}
	= \textbf{\emph{R}}
	\begin{pmatrix}
	x_n \\
	y_n \\
	z_n \\
	\end{pmatrix}
	,
\end{equation}

\vspace{-1mm}

with the rotation matrix $\textbf{\textit{R}}$. The displacement $\Delta \tilde{x}_n, \Delta \tilde{y}_n, \Delta \tilde{z}_n$ is the difference between the transformed coordinate systems. The displacement in global coordinates is obtained by a back-transformation of the predicted displacement, such that

\vspace{-3mm}

\begin{equation}\label{equ_backtransformation3}
	\textbf{\emph{R}}^{T} = \textbf{\emph{R}}^{-1} \; \; \text{and} \; \; \textbf{\emph{R}}^{T}\textbf{\emph{R}} = \textbf{\emph{R}}\textbf{\emph{R}}^{T} = \textbf{\emph{I}}.
\end{equation}

\vspace{-1mm}

\begin{figure}[!b]%
    \centering%
    \vspace{-3mm}
    \includegraphics[width=1\linewidth]{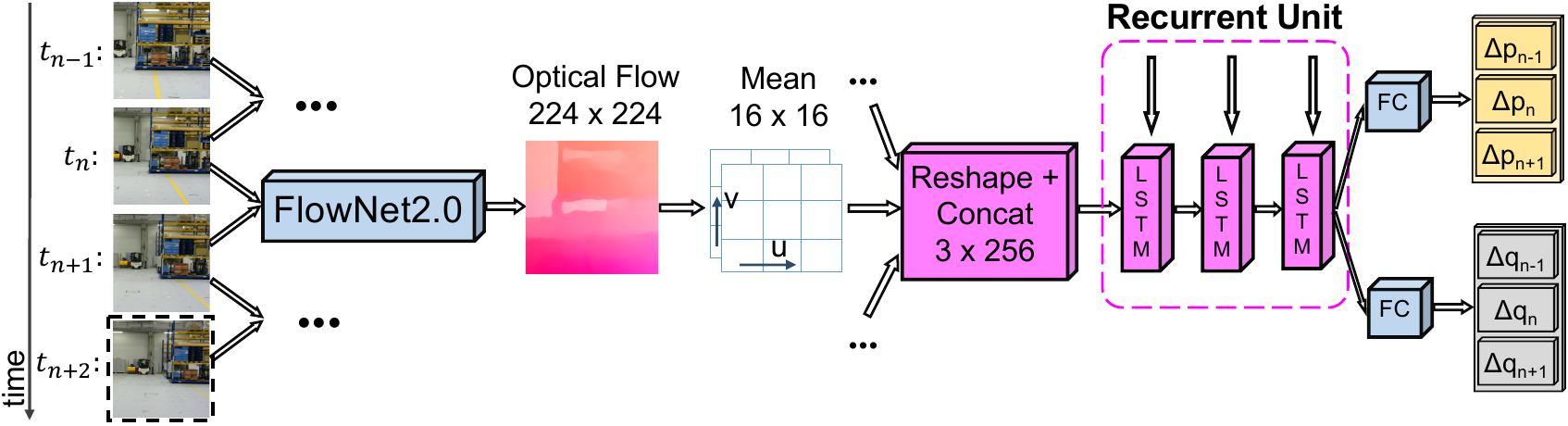}%
    \vspace{-2mm}
    \caption{Pipeline of the \textit{relative} pose regression (RPR) architecture: Data preprocessing, OF- and mean computation, reshaping, and concatenation, 3 recurrent LSTM units, and 2 FC-layers that yield the relative pose.}%
    \label{image_relative_reloc}%
\end{figure}

Fig.~\ref{image_relative_reloc} shows the structure of the RPR-network. Similar to the APR-network, the RPR-network also uses a stack of images, i.e., three OF-fields from the four input images of the timesteps $t_{n-1}, \dots, t_{n+2}$, to include more time context.

In a preliminary study, we found that our recurrent units struggle to remember temporal features when the direct input of the OF is too large (raw size $224 \times 224 \times 3$\,\textit{px}). This is in line with findings from Walch et al.~\cite{walch_lstm}. Hence, we split the OF in zones and compute the mean value for each the $u$- and $v$-direction. We reshape $16 \times 16$ number of zones in both directions to the size $2 \times 256$. The final concatenation results in a smaller total size of $3 \times 512$. The LSTM-output is forwarded to 2 FC-layers that regress both the displacement (size $3 \times 3$) and rotation (size $3 \times 4$).

The 2 FC-layers use the following loss function to predict the relative transposed poses $\Delta \tilde{\textbf{\textit{p}}}^{tr}$ and $\Delta \textbf{\emph{q}}$:

\vspace{-0.5cm}

\begin{equation}\label{equ_min_func_1}%
	\mathcal{L} = \alpha_2 \left\lVert{\Delta \tilde{\textbf{\emph{p}}}^{tr} - \Delta \textbf{\emph{p}}^{tr}}\right\rVert_2 + \beta_2 \left\lVert{\Delta \tilde{\textbf{\emph{q}}} - \frac{\Delta \textbf{\emph{q}}}{\left\lVert{\Delta \textbf{\emph{q}}}\right\rVert_2}}\right\rVert_2.
\end{equation}%

\vspace{-0.2cm}

The first term accounts for the predicted and transformed displacement $\Delta \tilde{\textbf{\textit{p}}}^{tr}$ to the ground truth displacement $\Delta \textbf{\textit{p}}^{tr}$ with an $L^2$-norm. The second term quantifies the error of the predicted rotation to the normalized ground truth rotation using an $L^2$-norm. Both terms are weighted by the hyperparameters $\alpha_2$ and $\beta_2$. A preliminary grid search with a fixed $\alpha_2=1$ revealed an optimal value for $\beta_2$ that depends on the scaling of the environment.

\subsection{6DoF Pose Estimation (PE) Network}\label{chapter_model_fusenet}%

\begin{figure}[t!]%
    \centering%
    \includegraphics[width=1\linewidth]{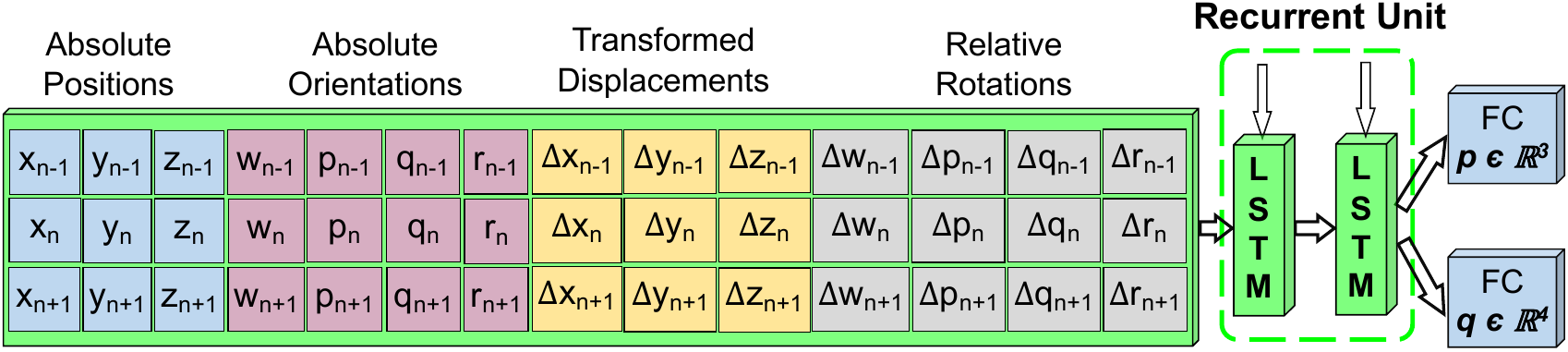}%
    \caption{Pipeline of the 6DoF PE-architecture. The input tensor ($3 \times 14$) contains absolute positions and orientations and relative displacements and rotations at timesteps $t_{n-1}, t_n, t_{n+1}$. 2 stacked LSTMs process the tensor and 2 FC-layers return the pose.}%
    \vspace{-0.6cm}%
    \label{image_lstm_input}%
\end{figure}

Our PE-network predicts absolute 6DoF poses from the outputs of both the APR- and RPR-networks, see Fig.~\ref{image_lstm_input}. The PE-network takes as input the absolute position $\textbf{\textit{p}}_{i} = (x_{i}, y_{i}, z_{i})$, the absolute orientation $\textbf{\textit{q}}_{i} = (w_{i}, p_{i}, q_{i}, r_{i})$, the relative displacement $\Delta \textbf{\textit{p}}_{i} = (\Delta x_{i}, \Delta y_{i}, \Delta z_{i})$, and the rotation change $\Delta \textbf{\textit{q}}_{i} = (\Delta w_{i}, \Delta p_{i}, \Delta q_{i}, \Delta r_{i})$. As we feed poses from three sequential timesteps $t_{n-1}$, $t_{n}$, and $t_{n+1}$ as input to the model it is implicitly \textit{time-distributed}. The 2 stacked LSTM-layers and the 2 FC-layers return a 3DoF absolute position $\textbf{\textit{p}} \in \mathbb{R}^{3}$ and a 3DoF orientation $\textbf{\textit{q}} \in \mathbb{R}^{4}$ using the following loss:

\vspace{-0.5cm}

\begin{equation}\label{equ_min_func_3}
	\mathcal{L}(P,\Delta P) = \alpha_3 \left\lVert{\tilde{\textbf{\emph{p}}} - \textbf{\emph{p}}}\right\rVert_2 + \beta_3 \left\lVert{\tilde{\textbf{\emph{q}}} - \frac{\textbf{\emph{q}}}{\left\lVert{ \textbf{\emph{q}}}\right\rVert_2}}\right\rVert_2.
\end{equation}


Again, in a preliminary grid search we chose $L_{2}$-norms with a fixed $\beta_3 = 1$ that revealed an optimal value for $\alpha_3$.

\section{Evaluation Datasets}\label{chapter_datasets}%

To train our network we need two different types of image data: (1) images annotated with their absolute poses for the APR-network, and (2) images of OF, annotated with their relative poses for the RPR-network.%

\textbf{Datasets to evaluate APR.} Publicly available datasets for absolute pose regression (\texttt{Cambridge Landmarks}~\cite{kendall_pose} and \texttt{TUM-LSI}~\cite{walch_lstm}) either lack accurate ground truth labels or the proximity between consecutive images is too large to embed meaningful temporal context. The \texttt{Aalto University}~\cite{laskar}, \texttt{Oxford RobotCar}~\cite{robotcar}, \texttt{DeepLoc}~\cite{vlocnet2} and \texttt{CMU Seasons}~\cite{sattler_cmu} datasets solve the small-scale issue of the \texttt{7-Scenes}~\cite{shotton_scene} dataset, but are barely used for evaluation of state-of-the-art techniques or consider only automotive-driving scenarios. The \texttt{12-Scenes}~\cite{12scenes} dataset is only used by \texttt{DSAC++}~\cite{dsac_pp}. For our industrial application these datasets are insufficient. \texttt{7-Scenes}~\cite{shotton_scene} only embeds scenes with less training data and only enables small scene-wise evaluations, but is mainly used for evaluation.
Hence, to compare ViPR with recent techniques we use the \texttt{7-Scenes}~\cite{shotton_scene} dataset. Furthermore, we recorded the \textit{Industry} dataset (see Sec.~\ref{chapter_datasets_novel}) that embeds three different industrial-like scenarios to allow a comprehensive and detailed evaluation with different movement patterns (such as slow motion and fast rotation).

\textbf{Datasets to evaluate RPR.} To evaluate the performance of the RPR and its contribution to ViPR, we also need a dataset with a close proximity between consecutive images. This is key to calculate the relative movement with OF. However, most publicly available datasets (\texttt{Middlebury}~\cite{baker_dataset}, \texttt{MPI Sintel}~\cite{butler_sintel}, \texttt{KITTI Vision}~\cite{geiger_kitti}, and \texttt{FlyingChairs}~\cite{flownet}) either do not meet this requirement or the OF pixel velocities do not match those of real-world applications. Hence, we directly calculate the OF from images with \texttt{FlowNet2.0}~\cite{flownet2} to train the RPR on it. Our novel \textit{Industry} dataset allows this, while retaining a large, diverse environment with hard real-world conditions, as described in the following.%

\subsection{Industry Dataset}\label{chapter_datasets_novel}%

\begin{figure*}[!t]
    \centering
    \begin{minipage}[h]{0.25\linewidth}
        \centering
        \includegraphics[height=1.6cm]{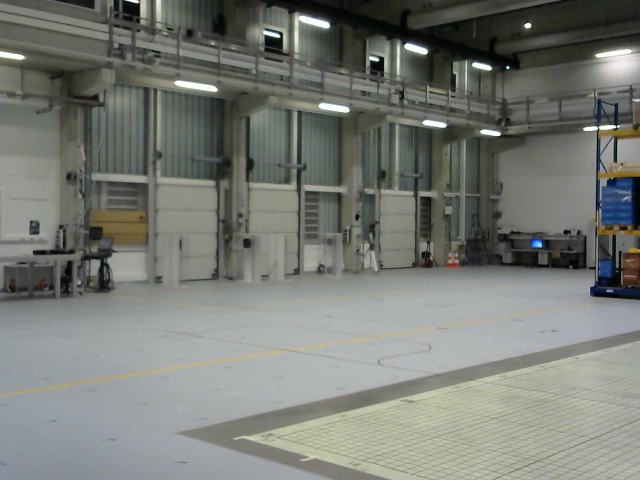}
        \includegraphics[height=1.6cm]{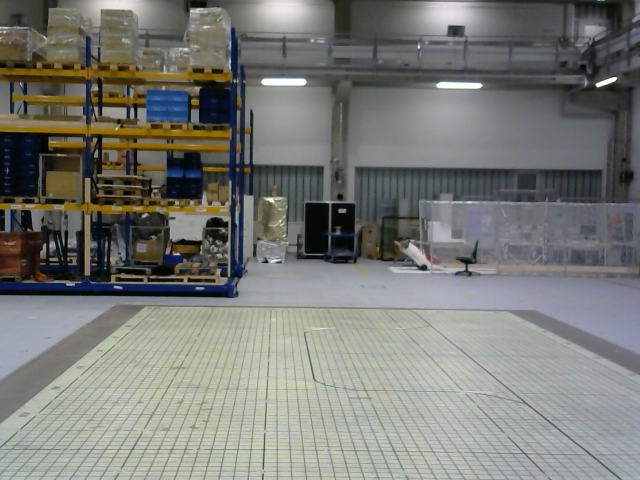}
        \subcaption{Scenario \#1 example images.}
        \label{image_scenario1}
    \end{minipage}
    \hspace{2.5mm}
    \begin{minipage}[h]{0.35\linewidth}
        \centering
        \includegraphics[height=1.6cm]{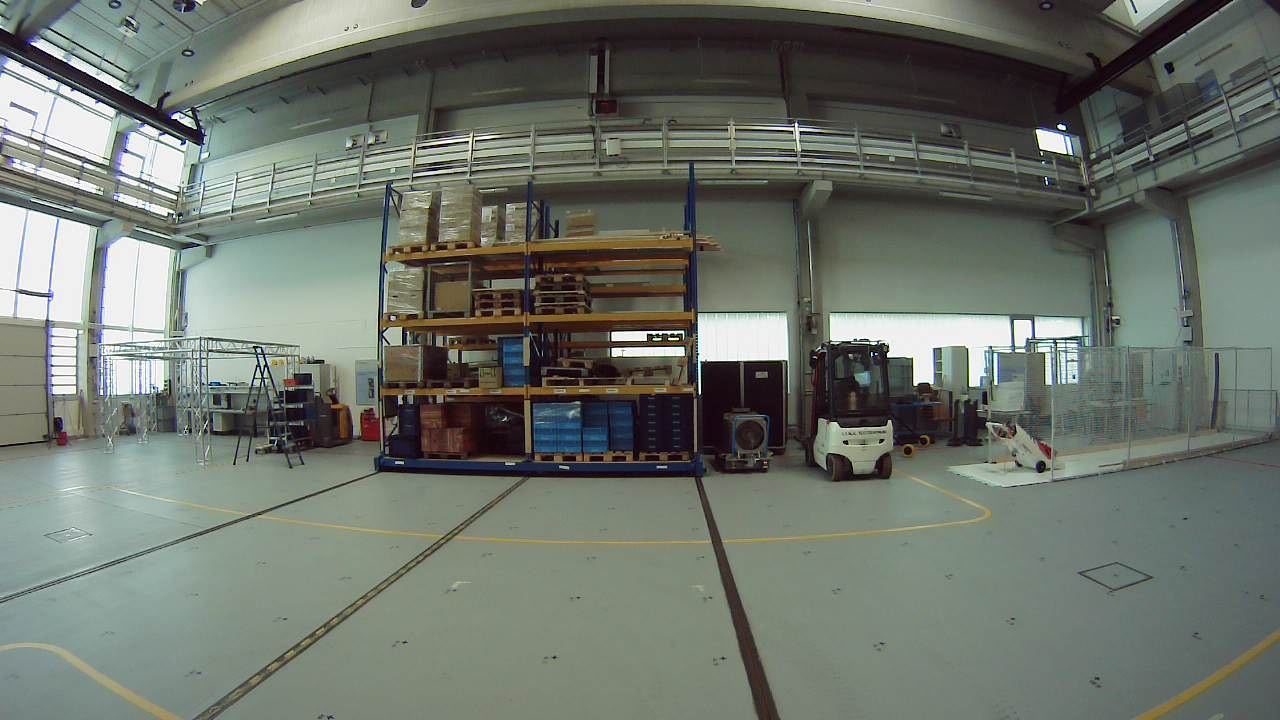}
        \includegraphics[height=1.6cm]{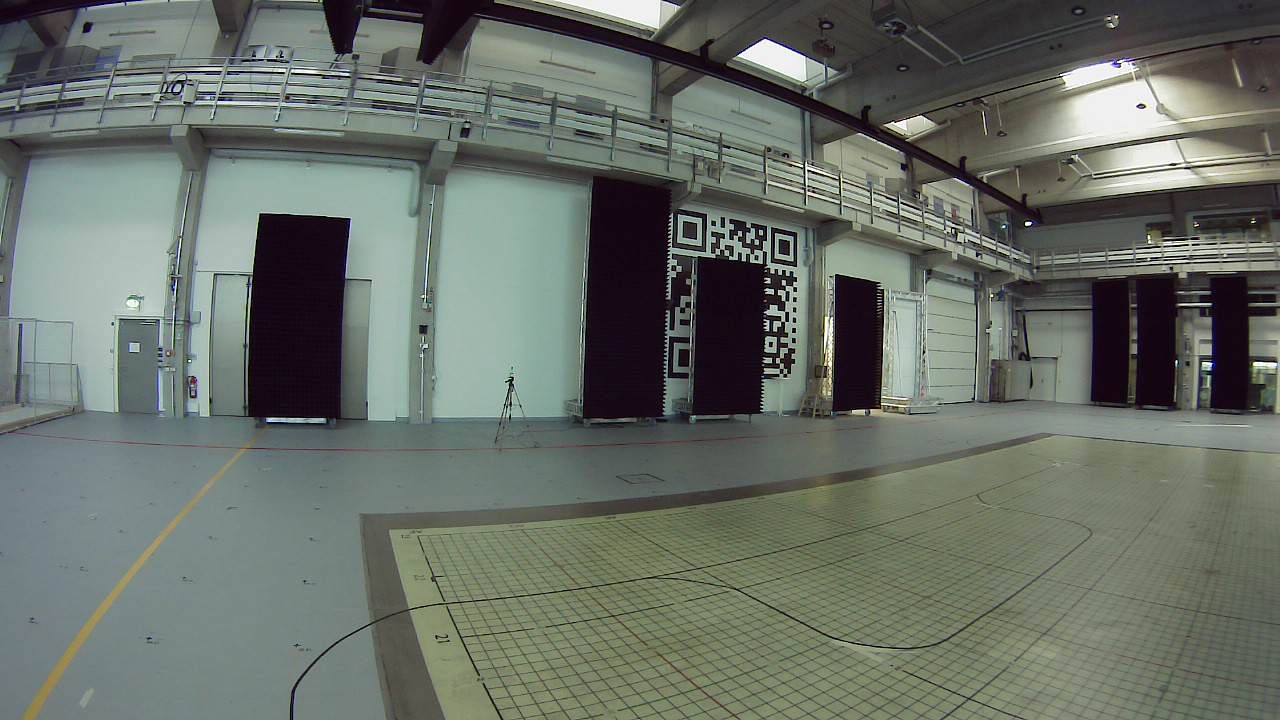}
        \subcaption{Scenario \#2 example images.}
        \label{image_scenario2}
    \end{minipage}
    \hspace{2.5mm}
    \begin{minipage}[h]{0.35\linewidth}
        \centering
        \includegraphics[height=1.6cm]{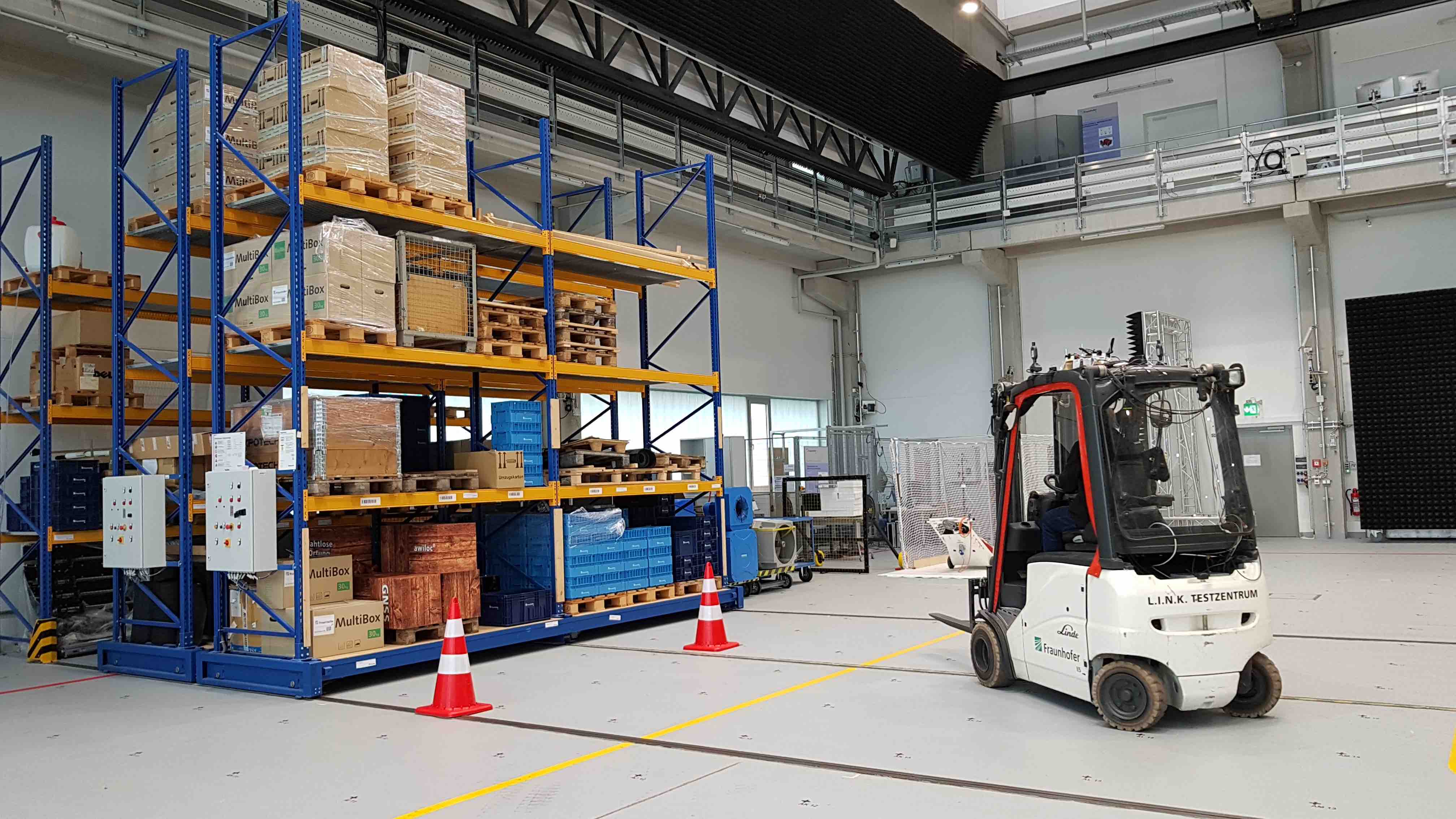}
        \includegraphics[height=1.6cm]{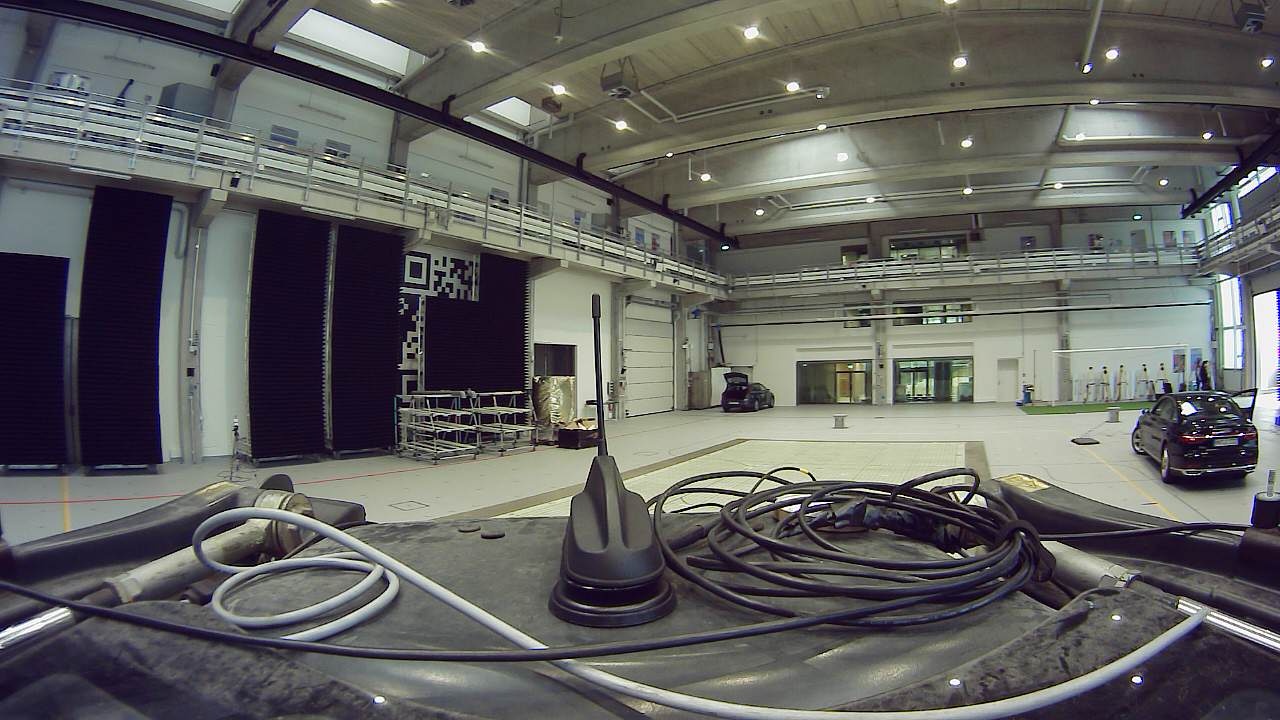}
        \subcaption{Scenario \#3 setup and example image.}
        \label{image_scenario3}
    \end{minipage}
    \vspace{-0.25cm}
    \caption{\textbf{Industry datasets.} Setup of the measurement environment (i.e., forklift truck, warehouse racks and black walls) and example images with normal (a) and wide-angle (b+c) cameras.}
    \vspace{-2mm}%
    \label{image_jetson_wideangle_3}
\end{figure*}

We designed the \textit{Industry} dataset to suite the requirements of both the APR- and the RPR-network and published the data\footnote{\textit{Industry} dataset available at: \href{https://www.iis.fraunhofer.de/de/ff/lv/lok/tech/opt1/warehouse.html}{https://www.iis.fraunhofer.de/warehouse}. \newline Provided are \textbf{raw images} and corresponding labels: $\textbf{\textit{p}}$ and $\textbf{\textit{q}}$.} at large-scale ($1,320m^2$) using a high-precision ($<1mm$) laser-based reference system. Each scenario presents different challenges (such as dynamic ego-motion with motion blur), various environmental characteristics (such as different geometric scales, light changes, i.e., artificial and natural light), and ambiguously structured elements, see Fig.~\ref{image_jetson_wideangle_3}.%

\begin{figure}[b!]
    \vspace{-3mm}
    \begin{minipage}[t]{0.49\linewidth}
    \begin{minipage}[t]{0.49\linewidth}
        \centering
        \hspace{-2mm}
        \includegraphics[width=1.05\linewidth]{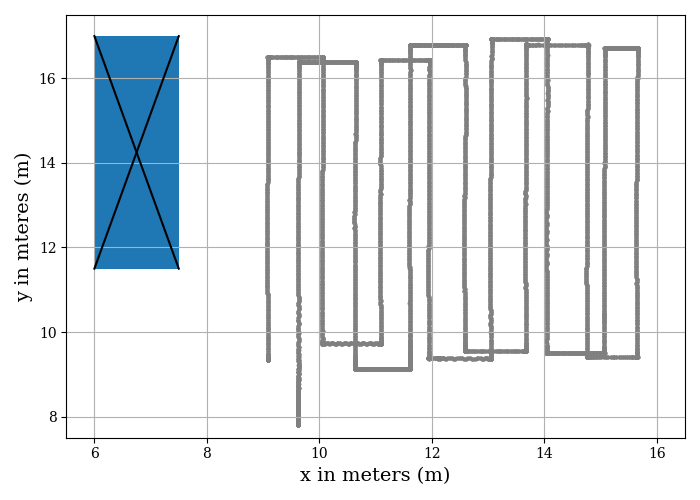}
        \subcaption{Training.}
        \label{dataset_3_angles_train}
    \end{minipage}
    \hfill
    \begin{minipage}[t]{0.49\linewidth}
        \centering
        \includegraphics[trim=12 0 10 0, clip, width=1\linewidth]{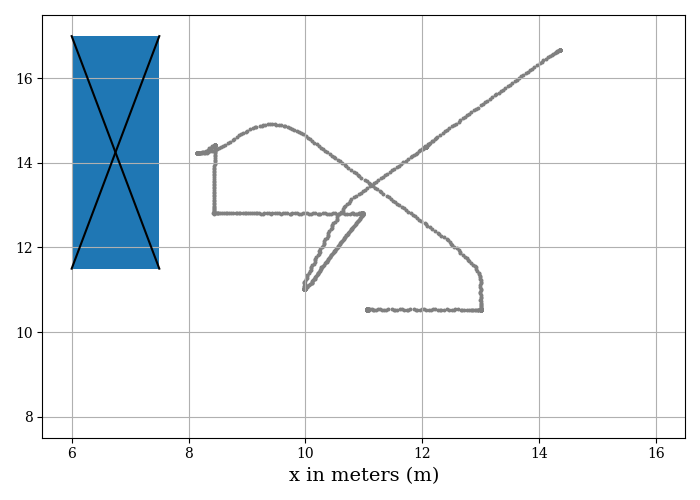}
        \subcaption{Testing.}
        \label{dataset_3_angles_test}
    \end{minipage}
    \label{dataset_3_angles}
    \end{minipage}
    \hfill
    \begin{minipage}[t]{0.49\linewidth}
    \begin{minipage}[t]{0.49\linewidth}
        \centering
        \includegraphics[trim=12 0 10 0, clip,width=1\linewidth]{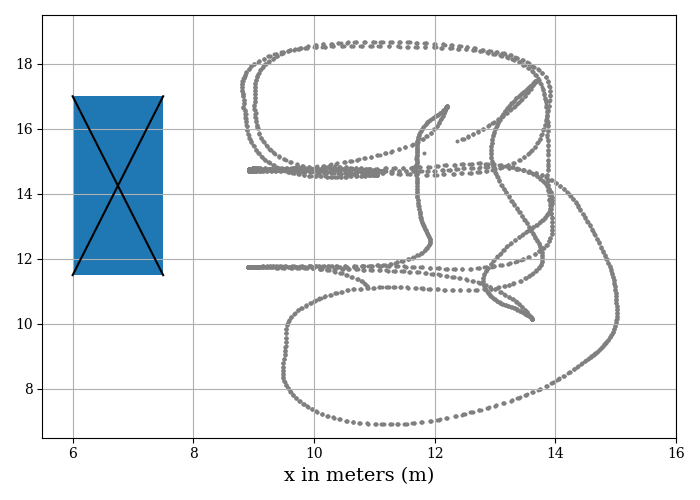}
        \subcaption{Training.}
        \label{dataset_4_angles_train}
    \end{minipage}
    \hfill
    \begin{minipage}[t]{0.49\linewidth}
        \centering
        \includegraphics[trim=12 0 10 0, clip,width=1\linewidth]{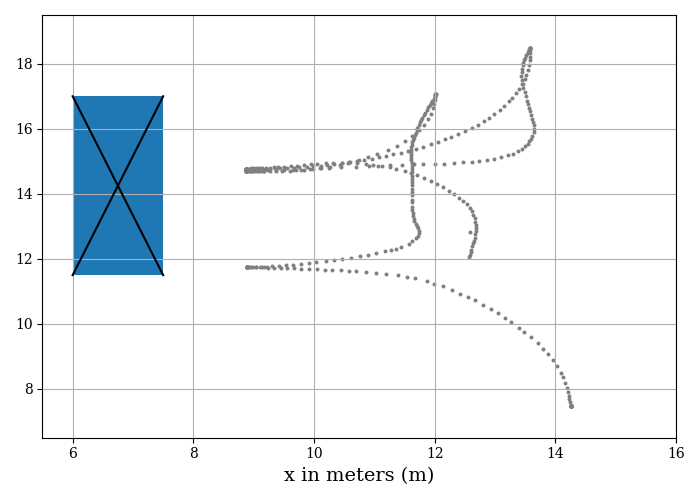}
        \subcaption{Testing.}
        \label{dataset_4_angles_test}
    \end{minipage}
    \label{dataset_4_angles}
    \end{minipage}
    \vspace{-0.3cm}
    \caption{Exemplary trajectories of \textit{Industry Scenarios \#2} (a-b) and \textit{\#3} (c-d) to assess the generalizability of ViPR.}
    \label{image_trajectory}
\end{figure}

\textbf{Industry Scenario \#1}~\cite{loeffler} has been recorded with 8 cameras (approx. $\ang{60}$ field-of-view (FoV) each) mounted on a stable apparatus to cover $\ang{360}$ (with overlaps) that has been moved automatically at a constant velocity of approx. 0.3\,$m/s$. The height of the cameras is at 1.7\,\textit{m}. The scenario contains 521,256 images ($640 \times 480$\,\textit{px}) and densely covers an area of 1,320\,\textit{m\textsuperscript{2}}. The environment imitates a typical warehouse scenario under realistic conditions. Besides well-structured elements such as high-level racks with goods, there are also very ambiguous and homogeneously textured elements (e.g., blank white or dark black walls). Both natural and artificial light illuminates volatile structures such as mobile work benches. While the training dataset is composed of a horizontal and vertical zig-zag movement of the apparatus the test datasets movements vary to cover different properties for a detailed evaluation, e.g., different environmental scalings (i.e., \textit{scale transition}, \textit{cross}, \textit{large scale}, and \textit{small scale}), network generalization (i.e., \textit{generalize open}, \textit{generalize racks}, and \textit{cross}), fast rotations (i.e., \textit{motion artifacts} was recorded on a forklift at 2.26\,\textit{m} height) and volatile objects (i.e., \textit{volatility}).

\textbf{Industry Scenario \#2} uses three $\ang{170}$ cameras (with overlaps) on the same apparatus at the same height. The recorded 11,859 training images ($1,280 \times 720$\,\textit{px}) represent a horizontal zig-zag movement (see Fig.~\ref{image_trajectory}a) and 3,096 test images represent a diagonal movement (see Fig.~\ref{image_trajectory}b). Compared to Scenario \#1 this scenario has more variation in its velocities (between 0\,$m/s$ and 0.3\,$m/s$, SD 0.05\,$m/s$).%

\textbf{Industry Scenario \#3} uses four $\ang{170}$ cameras (with overlaps) on a forklift truck at a height of 2.26\,$m$. Both the training and test datasets represents camera movements at varying, faster, and dynamic speeds (between 0\,$m/s$ and 1.5\,$m/s$, SD 0.51\,$m/s$). This makes the scenario the most challenging one. The training trajectory (see Fig.~\ref{image_trajectory}c) consists of 4,166 images and the test trajectory (see Fig.~\ref{image_trajectory}d) consists of 1,687 images. In contrast to the Scenarios \#1 and \#2 we train and test a typical industry scenario on dynamic movements of a forklift truck. However, one of cameras' images were corrupted in the test dataset, and thus, not used in the evaluation.
\section{Experimental Results}\label{chapter_results}%

\newcommand\tabrotate[1]{\rotatebox{90}{#1\hspace{\tabcolsep}}}
\begin{table*}
\begin{center}
\captionof{table}{Pose estimation results (position and orientation median error in meters $m$ and degrees ($^{\circ}$)) and total improvement of PE in $\%$ on the \texttt{7-Scenes}~\cite{shotton_scene} and \textit{Industry} datasets. The best results are bold and underlined ones are additionally referenced in the text.}
\vspace{-0.2cm}
\label{table_evaluation}
\footnotesize \begin{tabular}{ p{0.1cm} p{0.1cm} p{0.8cm} | p{1.2cm} | p{0.1cm} p{0.1cm} | p{0.1cm} p{0.1cm} | p{0.1cm} p{0.1cm} | p{0.1cm} p{0.1cm} | p{0.1cm} p{0.1cm} | p{1.4cm} }

	\multicolumn{3}{c|}{Dataset} & \multicolumn{1}{c|}{Spatial} & \multicolumn{2}{c|}{PoseNet~\cite{kendall_pose}} & \multicolumn{2}{c||}{PoseNet+} & \multicolumn{2}{c|}{APR-only} & \multicolumn{2}{c|}{APR+LSTM} & \multicolumn{2}{c|}{ViPR*} & \multicolumn{1}{c}{Improv.} \\
	
	& & & \multicolumn{1}{c|}{extend ($m$)} & \multicolumn{2}{c|}{(\textit{original}/our param.)} & \multicolumn{2}{c||}{LSTM~\cite{walch_lstm}} & \multicolumn{2}{c|}{\textit{}} & \multicolumn{2}{c|}{(our param.)} & & & \multicolumn{1}{c}{ViPR (\%)} \\ \hline
	
	& \multicolumn{2}{l|}{chess} & \multicolumn{1}{r|}{$3.0\!\times\!2.0\!\times\!1.0$} & \multicolumn{1}{r}{\textit{0.32}\,\slash\,0.24} & \multicolumn{1}{r|}{\textit{4.06}\,\slash\,7.79} & \multicolumn{1}{r}{\textit{0.24}} & \multicolumn{1}{r||}{\textit{5.77}} & \multicolumn{1}{r}{0.23} & \multicolumn{1}{r|}{7.96} & \multicolumn{1}{r}{0.27} & \multicolumn{1}{r|}{9.66} & \multicolumn{1}{r}{\textbf{0.22}} & \multicolumn{1}{r|}{\textbf{7.89}} & \multicolumn{1}{r}{\textbf{+}\,1.74} \\
	
    \multirow{4}{*}{\tabrotate{\texttt{7-Scenes} \cite{shotton_scene}}} & \multicolumn{2}{l|}{fire} & \multicolumn{1}{r|}{$2.5\!\times\!1.0\!\times\!1.0$} & \multicolumn{1}{r}{\textit{0.47}\,\slash\,0.39} & \multicolumn{1}{r|}{\textit{14.4}\,\slash\,12.40} & \multicolumn{1}{r}{\textit{0.34}} & \multicolumn{1}{r||}{\textit{11.9}} & \multicolumn{1}{r}{0.39} & \multicolumn{1}{r|}{12.85} & \multicolumn{1}{r}{0.50} & \multicolumn{1}{r|}{15.70} & \multicolumn{1}{r}{\textbf{0.38}} & \multicolumn{1}{r|}{\textbf{12.74}} & \multicolumn{1}{r}{\textbf{+}\,2.56} \\
    
    & \multicolumn{2}{l|}{heads} & \multicolumn{1}{r|}{$2.0\!\times\!0.5\!\times\!1.0$} & \multicolumn{1}{r}{\textit{0.29}\,\slash\,0.21} & \multicolumn{1}{r|}{\textit{6.00}\,\slash\,16.46} & \multicolumn{1}{r}{\textit{0.21}} & \multicolumn{1}{r||}{\textit{13.7}} & \multicolumn{1}{r}{0.22} & \multicolumn{1}{r|}{16.48} & \multicolumn{1}{r}{0.23} & \multicolumn{1}{r|}{16.91} & \multicolumn{1}{r}{\textbf{0.21}} & \multicolumn{1}{r|}{\textbf{16.41}} & \multicolumn{1}{r}{\textbf{+}\,3.64} \\
	
    & \multicolumn{2}{l|}{office} & \multicolumn{1}{r|}{$2.5\!\times\!2.0\!\times\!1.5$} & \multicolumn{1}{r}{\textit{0.48}\,\slash\,0.33} & \multicolumn{1}{r|}{\textit{3.84}\,\slash\,10.08} & \multicolumn{1}{r}{\textit{0.30}} & \multicolumn{1}{r||}{\textit{8.08}} & \multicolumn{1}{r}{0.36} & \multicolumn{1}{r|}{10.11} & \multicolumn{1}{r}{0.37} & \multicolumn{1}{r|}{10.83} & \multicolumn{1}{r}{\textbf{0.35}} & \multicolumn{1}{r|}{\textbf{9.59}} & \multicolumn{1}{r}{\textbf{+}\,4.01} \\
	
    & \multicolumn{2}{l|}{pumpkin} & \multicolumn{1}{r|}{$2.5\!\times\!1.0\!\times\!1.0$} & \multicolumn{1}{r}{\textit{0.47}\,\slash\,0.45} & \multicolumn{1}{r|}{\textit{8.42}\,\slash\,8.70} & \multicolumn{1}{r}{\textit{0.33}} & \multicolumn{1}{r||}{\textit{7.00}} & \multicolumn{1}{r}{0.39} & \multicolumn{1}{r|}{8.57} & \multicolumn{1}{r}{0.86} & \multicolumn{1}{r|}{49.46} & \multicolumn{1}{r}{\textbf{0.37}} & \multicolumn{1}{r|}{\textbf{8.45}} & \multicolumn{1}{r}{\textbf{+}\,5.12} \\
    
    & \multicolumn{2}{l|}{red kitchen} & \multicolumn{1}{r|}{$4.0\!\times\!3.0\!\times\!1.5$} & \multicolumn{1}{r}{\textit{0.59}\,\slash\,0.41} & \multicolumn{1}{r|}{\textit{8.64}\,\slash\,9.08} & \multicolumn{1}{r}{\textit{0.37}} & \multicolumn{1}{r||}{\textit{8.83}} & \multicolumn{1}{r}{0.42} & \multicolumn{1}{r|}{9.33} & \multicolumn{1}{r}{1.06} & \multicolumn{1}{r|}{50.67} & \multicolumn{1}{r}{\textbf{0.40}} & \multicolumn{1}{r|}{\textbf{9.32}} & \multicolumn{1}{r}{\textbf{+}\,4.76} \\
    
    & \multicolumn{2}{l|}{stairs} & \multicolumn{1}{r|}{$2.5\!\times\!2.0\!\times\!1.5$} & \multicolumn{1}{r}{\textit{0.47}\,\slash\,0.36} & \multicolumn{1}{r|}{\textit{6.93}\,\slash\,13.69} & \multicolumn{1}{r}{\textit{0.40}} & \multicolumn{1}{r||}{\textit{13.7}} & \multicolumn{1}{r}{0.31} & \multicolumn{1}{r|}{12.49} & \multicolumn{1}{r}{0.42} & \multicolumn{1}{r|}{13.50} & \multicolumn{1}{r}{\textbf{0.31}} & \multicolumn{1}{r|}{12.65} & \multicolumn{1}{r}{\textbf{+}\,0.46} \\ \cline{2-15}
    
    & \multicolumn{2}{l|}{$\diameter$ total} & & \multicolumn{1}{r}{\textit{0.44}\,\slash\,\underline{0.34}} & \multicolumn{1}{r|}{\textit{7.47}\,\slash\,11.17} & \multicolumn{1}{r}{\textit{0.31}} & \multicolumn{1}{r||}{\textit{9.85}} & \multicolumn{1}{r}{\underline{0.33}} & \multicolumn{1}{r|}{11.11} & \multicolumn{1}{r}{0.53} & \multicolumn{1}{r|}{23.82} & \multicolumn{1}{r}{\underline{\textbf{0.32}}} & \multicolumn{1}{r|}{\textbf{11.01}} & \multicolumn{1}{r}{\underline{\textbf{+\,3.18}}} \\ \hline \hline
    
    \multirow{4}{*}{\tabrotate{Industry \textit{Scenario 1} \cite{loeffler}}} & \multicolumn{2}{l|}{cross} & \multicolumn{1}{r|}{$24.5\!\times\!16.0$} & \multicolumn{1}{r}{--\,\slash\,1.15} & \multicolumn{1}{r|}{--\,\slash\,0.75} & \multicolumn{2}{c||}{--} & \multicolumn{1}{r}{0.61} & \multicolumn{1}{r|}{0.53} & \multicolumn{1}{r}{4.42} & \multicolumn{1}{r|}{0.21} & \multicolumn{1}{r}{\textbf{0.46}} & \multicolumn{1}{r|}{0.60} & \multicolumn{1}{r}{\underline{\textbf{+}\,25.31}} \\
    
    & \multicolumn{2}{l|}{gener. open} & \multicolumn{1}{r|}{$20.0\!\times\!17.0$} & \multicolumn{1}{r}{--\,\slash\,1.94} & \multicolumn{1}{r|}{--\,\slash\,11.73} & \multicolumn{2}{c||}{--} & \multicolumn{1}{r}{1.68} & \multicolumn{1}{r|}{11.07} &  \multicolumn{1}{r}{3.36} & \multicolumn{1}{r|}{2.95} & \multicolumn{1}{r}{\textbf{1.48}} & \multicolumn{1}{r|}{\textbf{10.86}} & \multicolumn{1}{r}{\underline{\textbf{+}\,11.75}} \\
    
    & \multicolumn{2}{l|}{gener. racks} & \multicolumn{1}{r|}{$8.5\!\times\!18.5$} & \multicolumn{1}{r}{--\,\slash\,3.48} & \multicolumn{1}{r|}{--\,\slash\,6.01} & \multicolumn{2}{c||}{--} & \multicolumn{1}{r}{2.48} & \multicolumn{1}{r|}{1.53} &  \multicolumn{1}{r}{3.90} & \multicolumn{1}{r|}{0.61} & \multicolumn{1}{r}{\textbf{2.38}} & \multicolumn{1}{r|}{1.95} & \multicolumn{1}{r}{\underline{\textbf{+}\,4.03}} \\
    
    & \multicolumn{2}{l|}{large scale} & \multicolumn{1}{r|}{$19.0\!\times\!19.0$} & \multicolumn{1}{r}{--\,\slash\,2.32} & \multicolumn{1}{r|}{--\,\slash\,6.37} & \multicolumn{2}{c||}{--} & \multicolumn{1}{r}{2.37} & \multicolumn{1}{r|}{9.82} &  \multicolumn{1}{r}{4.99} & \multicolumn{1}{r|}{1.61} & \multicolumn{1}{r}{\textbf{2.12}} & \multicolumn{1}{r|}{8.64} & \multicolumn{1}{r}{\underline{\textbf{+}\,10.68}} \\
    
    & \multicolumn{2}{l|}{motion art.} & \multicolumn{1}{r|}{$37.0\!\times\!17.0$} & \multicolumn{1}{r}{--\,\slash\,7.43} & \multicolumn{1}{r|}{--\,\slash\,124.94} & \multicolumn{2}{c||}{--} & \multicolumn{1}{r}{7.48} & \multicolumn{1}{r|}{131.30} &  \multicolumn{1}{r}{8.18} & \multicolumn{1}{r|}{139.37} & \multicolumn{1}{r}{\textbf{6.73}} & \multicolumn{1}{r|}{136.6} & \multicolumn{1}{r}{\underline{\textbf{+}\,10.01}} \\
    
    & \multicolumn{2}{l|}{scale trans.} & \multicolumn{1}{r|}{$28.0\!\times\!19.5$} & \multicolumn{1}{r}{--\,\slash\,2.17} & \multicolumn{1}{r|}{--\,\slash\,3.03} & \multicolumn{2}{c||}{--} & \multicolumn{1}{r}{1.94} & \multicolumn{1}{r|}{6.46} &  \multicolumn{1}{r}{5.63} & \multicolumn{1}{r|}{0.58} & \multicolumn{1}{r}{\textbf{1.64}} & \multicolumn{1}{r|}{6.29} & \multicolumn{1}{r}{\underline{\textbf{+}\,15.52}} \\
    
    & \multicolumn{2}{l|}{small scale} & \multicolumn{1}{r|}{$10.0\!\times\!11.0$} & \multicolumn{1}{r}{--\,\slash\,3.78} & \multicolumn{1}{r|}{--\,\slash\,9.18} & \multicolumn{2}{c||}{--} & \multicolumn{1}{r}{4.09} & \multicolumn{1}{r|}{20.75} &  \multicolumn{1}{r}{4.46} & \multicolumn{1}{r|}{6.06} & \multicolumn{1}{r}{\textbf{3.50}} & \multicolumn{1}{r|}{15.74} & \multicolumn{1}{r}{\underline{\textbf{+}\,14.41}} \\
    
    & \multicolumn{2}{l|}{volatility} & \multicolumn{1}{r|}{$29.0\!\times\!13.0$} & \multicolumn{1}{r}{--\,\slash\,2.68} & \multicolumn{1}{r|}{--\,\slash\,78.52} & \multicolumn{2}{c||}{--} & \multicolumn{1}{r}{2.09} & \multicolumn{1}{r|}{77.68} &  \multicolumn{1}{r}{4.16} & \multicolumn{1}{r|}{78.73} & \multicolumn{1}{r}{\textbf{1.96}} & \multicolumn{1}{r|}{\underline{\textbf{77.54}}} & \multicolumn{1}{r}{\underline{\textbf{+}\,6.41}} \\ \cline{2-15}
    
    & \multicolumn{2}{l|}{$\diameter$ total} & & \multicolumn{1}{r}{--\,\slash\,3.12} & \multicolumn{1}{r|}{--\,\slash\,30.07} & \multicolumn{2}{c||}{--} & \multicolumn{1}{r}{2.82} & \multicolumn{1}{r|}{32.30} & \multicolumn{1}{r}{\underline{4.89}} & \multicolumn{1}{r|}{28.76} & \multicolumn{1}{r}{\underline{\textbf{2.53}}} & \multicolumn{1}{r|}{32.28} & \multicolumn{1}{r}{\underline{\textbf{+\,12.27}}} \\ \hline \hline
    
    \multirow{3}{*}{\tabrotate{Industry}} & \multirow{3}{*}{\tabrotate{\textit{Scen. 2}}} & \multicolumn{1}{l|}{cam \#0} & \multicolumn{1}{r|}{$6.5\!\times\!9.0$} & \multicolumn{1}{r}{--\,\slash\,0.49} & \multicolumn{1}{r|}{--\,\slash\,0.21} & \multicolumn{2}{c||}{--} & \multicolumn{1}{r}{0.22} & \multicolumn{1}{r|}{0.29} &  \multicolumn{1}{r}{1.49} & \multicolumn{1}{r|}{0.14} & \multicolumn{1}{r}{\textbf{0.16}} & \multicolumn{1}{r|}{3.37} & \multicolumn{1}{r}{\textbf{+}\,26.24} \\
    
    & & \multicolumn{1}{l|}{cam \#1} & \multicolumn{1}{r|}{$6.5\!\times\!9.0$} & \multicolumn{1}{r}{--\,\slash\,0.15} & \multicolumn{1}{r|}{--\,\slash\,0.38} & \multicolumn{2}{c||}{--} & \multicolumn{1}{r}{0.23} & \multicolumn{1}{r|}{0.35} &  \multicolumn{1}{r}{2.68} & \multicolumn{1}{r|}{0.17} & \multicolumn{1}{r}{\textbf{0.12}} & \multicolumn{1}{r|}{2.75} & \multicolumn{1}{r}{\underline{\textbf{+}\,46.49}} \\
    
    & & \multicolumn{1}{l|}{cam \#2} & \multicolumn{1}{r|}{$6.5\!\times\!9.0$} & \multicolumn{1}{r}{--\,\slash\,0.43} & \multicolumn{1}{r|}{--\,\slash\,0.19} & \multicolumn{2}{c||}{--} & \multicolumn{1}{r}{\underline{0.37}} & \multicolumn{1}{r|}{0.13} & \multicolumn{1}{r}{0.90} & \multicolumn{1}{r|}{0.15} &  \multicolumn{1}{r}{\underline{\textbf{0.30}}} & \multicolumn{1}{r|}{1.84} & \multicolumn{1}{r}{\underline{\textbf{+}\,17.87}} \\ \cline{3-15}
    
    & & \multicolumn{1}{l|}{$\diameter$ total} & & \multicolumn{1}{r}{--\,\slash\,\underline{0.36}} & \multicolumn{1}{r|}{--\,\slash\,0.26} & \multicolumn{2}{c||}{--} & \multicolumn{1}{r}{\underline{0.27}} & \multicolumn{1}{r|}{0.26} & \multicolumn{1}{r}{\underline{1.69}} & \multicolumn{1}{r|}{0.15} & \multicolumn{1}{r}{\underline{\textbf{0.20}}} & \multicolumn{1}{r|}{2.65} & \multicolumn{1}{r}{\underline{\textbf{+\,30.20}}} \\ \hline \hline
    
    \multirow{3}{*}{\tabrotate{Industry}} & \multirow{3}{*}{\tabrotate{\textit{Scen. 3}}} & \multicolumn{1}{l|}{cam \#0} & \multicolumn{1}{r|}{$6.0\!\times\!11.0$} & \multicolumn{1}{r}{--\,\slash\,0.41} & \multicolumn{1}{r|}{--\,\slash\,1.00} & \multicolumn{2}{c||}{--} & \multicolumn{1}{r}{0.34} & \multicolumn{1}{r|}{1.26} &  \multicolumn{1}{r}{0.72} & \multicolumn{1}{r|}{1.31} & \multicolumn{1}{r}{\textbf{0.27}} & \multicolumn{1}{r|}{1.43} & \multicolumn{1}{r}{\underline{\textbf{+}\,20.64}} \\
    
    & & \multicolumn{1}{l|}{cam \#1} & \multicolumn{1}{r|}{$6.0\!\times\!11.0$} & \multicolumn{1}{r}{--\,\slash\,0.32} & \multicolumn{1}{r|}{--\,\slash\,1.07} & \multicolumn{2}{c||}{--} & \multicolumn{1}{r}{0.26} & \multicolumn{1}{r|}{1.11} & \multicolumn{1}{r}{0.88} & \multicolumn{1}{r|}{1.27} & \multicolumn{1}{r}{\textbf{0.21}} & \multicolumn{1}{r|}{\textbf{1.06}} & \multicolumn{1}{r}{\underline{\textbf{+}\,20.13}} \\
    
    & & \multicolumn{1}{l|}{cam \#2} & \multicolumn{1}{r|}{$6.0\!\times\!11.0$} & \multicolumn{1}{r}{--\,\slash\,0.32} & \multicolumn{1}{r|}{--\,\slash\,1.60} & \multicolumn{2}{c||}{--} & \multicolumn{1}{r}{0.36} & \multicolumn{1}{r|}{1.62} & \multicolumn{1}{r}{0.72} & \multicolumn{1}{r|}{1.74} & \multicolumn{1}{r}{\textbf{0.32}} & \multicolumn{1}{r|}{\textbf{1.38}} & \multicolumn{1}{r}{\underline{\textbf{+}\,11.47}} \\ \cline{3-15}
    
    & & \multicolumn{1}{l|}{$\diameter$ total} &  & \multicolumn{1}{r}{--\,\slash\,0.35} & \multicolumn{1}{r|}{--\,\slash\,1.22} & \multicolumn{2}{c||}{--} & \multicolumn{1}{r}{0.32} & \multicolumn{1}{r|}{1.33} &  \multicolumn{1}{r}{0.77} & \multicolumn{1}{r|}{1.44} & \multicolumn{1}{r}{\textbf{0.27}} & \multicolumn{1}{r|}{1.29} & \multicolumn{1}{r}{\underline{\textbf{+\,17.41}}} \\
    
\end{tabular}
\end{center}
\vspace{-0.9cm}
\end{table*}

To compare ViPR with state-of-the-art results, we first briefly describe our parameterization of \texttt{PoseNet}~\cite{kendall_pose} and \texttt{PoseNet+LSTM}~\cite{walch_lstm} in Sec.~\ref{subsection:baselines}. Next, Sec.~\ref{chapter_absnet_fusenet} presents our results. We highlight the performance of ViPR's sub-networks (APR, APR+LSTM) individually, and investigate both the impact of RPR and PE on the final pose estimation accuracy of ViPR. Sec.~\ref{chapter_relnet_eval} shows results of the RPR-network. Finally, we discuss general findings and show runtimes of our models in Sec.~\ref{chapter_runtime}.

For all experiments we used an AMD Ryzen 7 2700 CPU 3.2\,\textit{GHz} equipped with one NVidia GeForce RTX 2070 with 8 GB GDDR6 VRAM. Tab.~\ref{table_evaluation} shows the median error of the position in $m$ and the orientation in \textit{degrees}. The second column reports the spatial extends of the datasets. The last column reports the improvement in position accuracy of ViPR (in \%) over APR-only.

\subsection{Baselines}
\label{subsection:baselines}

As a baseline we report the initially described results on \texttt{7-Scenes} of \texttt{PoseNet}~\cite{kendall_pose} and \texttt{PoseNet+LSTM}~\cite{walch_lstm} (\textit{in italic}). We further re-implemented the initial variant of \texttt{PoseNet} and trained it from scratch with $\alpha_1=1$, $\beta_1=30$ (thus optimizing for positional accuracy at the expense of orientation accuracy). Tab.~\ref{table_evaluation} (cols. 3 and 4) shows our implementation's results next to the initially reported ones (on \texttt{7-Scenes}). We see that (as expected) the results of the \texttt{PoseNet} implementations differ due to changed values for $\alpha_1$ and $\beta_1$ in our implementation.

\subsection{Evaluation of the ViPR-Network}
\label{chapter_absnet_fusenet}%

In the following, we evaluate our method in multiple scenarios with different distinct challenges for the pose estimation task. \texttt{7-Scenes} focuses on difficult motion blur conditions of typical human motion. We then use the \textit{Industry Scenario \#1} to investigate various challenges at a larger scale, but with mostly constant velocities. \textit{Industry Scenarios \#2} and \textit{\#3} then focus on dynamic, fast ego-motion of a moving forklift truck at large-scale.

\textbf{7-Scenes~\cite{shotton_scene}.} For both architectures (\texttt{PoseNet} and ViPR), we optimized $\beta$ to weight the impact of position and orientation such that it yields the smallest total median error. Both APR+LSTM and ViPR return a slightly lower pose estimation error of 0.33\,$m$ and 0.32\,$m$ than PoseNet+LSTM with 0.34\,$m$. ViPR yields an average improvement of the position accuracy of \underline{\SI{3.18}{\percent}} even in strong motion blur situations. The results indicate that ViPR relies on a plausible optical flow component to achieve performance that is superior to the baseline. In situations of negligible motion between frames the median only improves by 0.02\,$m$. However, the average accuracy gain still shows that ViPR performs \emph{en par} or better than the baselines.

\textbf{Stable motion evaluation.} For the \textit{Industry Scenario \#1} dataset, we train the models on the zig-zag trajectories, and test them on specific sub-trajectories with individual challenges, but at almost constant velocity. In total, ViPR improves the position accuracy by \underline{$12.27\%$} on average (min.: \SI{4.03}{\percent}; max.: \SI{25.31}{\percent}) while the orientation error is similar for most of the architectures and test sets.

In environments with volatile features, i.e., objects that are only present in the test dataset, we found that ViPR (with optical flow) is significantly (\SI{6.41}{\percent}) better compared to APR-only. However, the high angular error of \SI{77.54}{\degree} indicates an irrecoverable degeneration of the APR-part. In tests with different scaling of the environment, we think that ViPR learns an interpretation of relative and absolute position regression, that works both in small and large proximity to environmental features, as ViPR improves by \SI{15.52}{\percent} (scale trans.) and \SI{14.41}{\percent} (small scale) or \SI{10.68}{\percent} (large scale). When the test trajectories are located within areas that embed only few or no training samples (gener. racks and open), ViPR still improves over other methods with 4.03-11.75\,$\%$. The highly dynamic test on a forklift truck (motion artifacts) is exceptional here as only the test dataset contains dynamics and blur, and hence, challenges ViPR most. However, ViPR still improves by \SI{10.01}{\percent} over APR-only, despite the data dynamic's absolute novelty.

In summary, ViPR decreases the position median significantly by about 2.53\,$m$ than only APR+LSTM (4.89\,$m$). This and the other findings are strong indicators that the relative component RPR significantly supports the final pose estimation of ViPR.

\begin{figure}[b!]
    \vspace{-4mm}
    \begin{minipage}[t]{0.51\linewidth}
    \centering
    \includegraphics[trim=0 0 0 0, clip, width=1.0\linewidth]{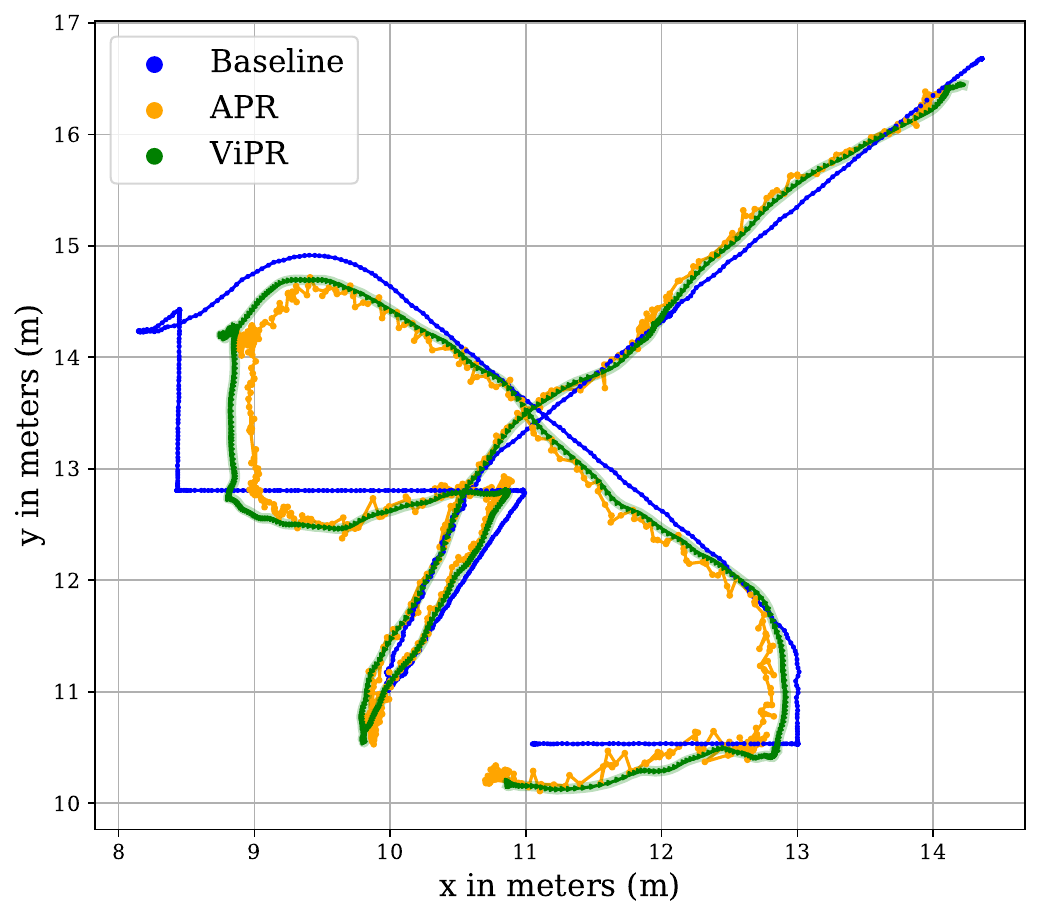}
    \subcaption{\textit{Scenario \#2}.}
    \label{image_absnet_kalman}
    \end{minipage}
    \hspace{-0.2cm}
    \begin{minipage}[t]{0.49\linewidth}
    \centering
    \includegraphics[trim=20 0 0 0, clip,width=1.0\linewidth]{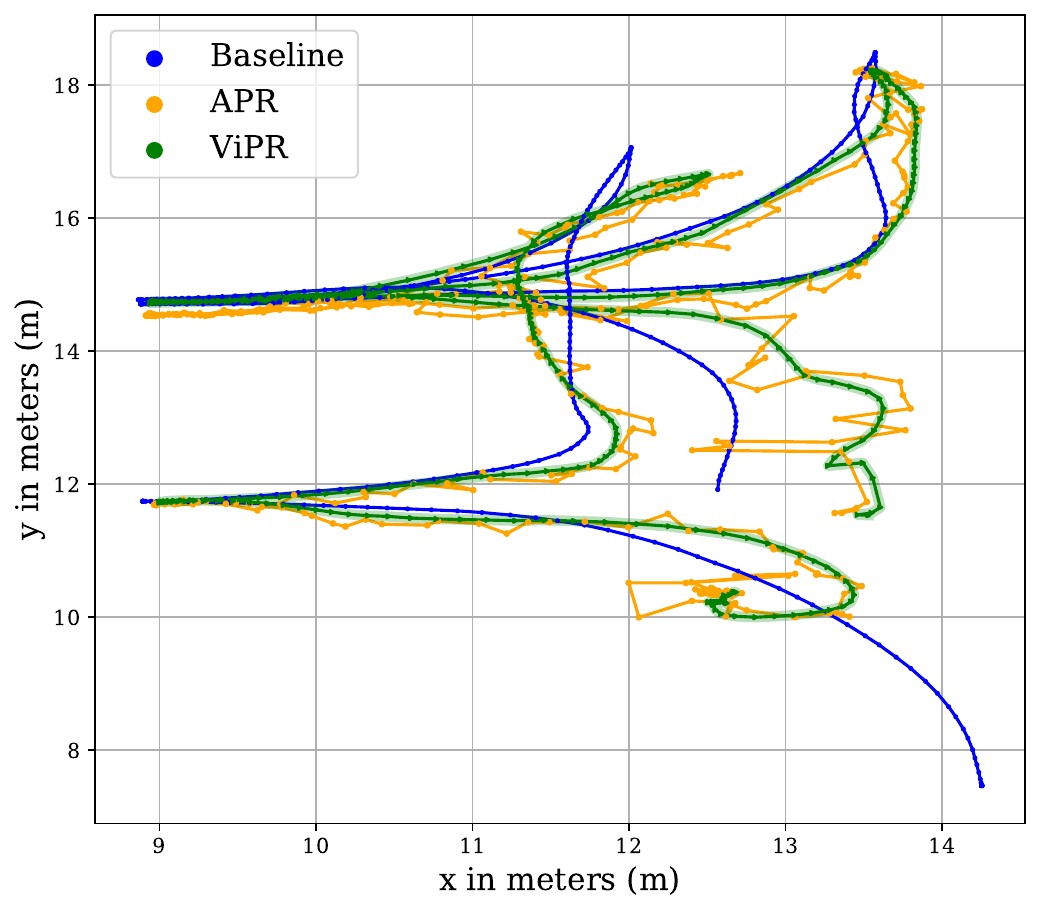}
    \subcaption{\textit{Scenario \#3}.}
    \label{image_com_pose_fuse_18}
    \end{minipage}
    \vspace{-2.5mm}
    \caption{Exemplary comparison of {\textcolor{orange}{APR}}, {\textcolor{green}{ViPR}}, and a {\textcolor{blue}{baseline}} (ground truth) trajectory of the \textit{Industry} datasets.}
    \label{image_apr_vipr}
\end{figure}

\textit{Industry Scenario \#2} is designed to evaluate for unknown trajectories. Hence, training trajectories represent an orthogonal grid, and test trajectories are diagonal. In total, ViPR improves the position accuracy by \underline{\SI{30.2}{\percent}} on average (min.: \SI{17.87}{\percent}; max.: \SI{46.49}{\percent}). Surprisingly, the orientation error is comparable for all architectures, except ViPR. We think that this is because ViPR learns to optimize its position based on the APR- and RPR- orientations, and hence, exploits these orientations to improve its position estimate, that we prioritized in the loss function. APR-only yields an average position accuracy of 0.27\,$m$, while the pure \texttt{PoseNet} yields position errors of 0.36\,$m$ on average, but APR+LSTM results in an even worse accuracy of 1.69\,$m$. Instead, the novel ViPR outperforms all significantly with 0.2\,$m$. Compared to our APR+LSTM approach, we think that ViPR on the one hand interprets and compensates the (long-term) drift of RPR and on the other hand smooths the short-term errors of APR, as PE counteracts the accumulation of RPR's scaling errors with APR's absolute estimates. Here, the synergies of the networks in ViPR are particularly effective. This is also visualized in Fig.~\ref{image_absnet_kalman}: the green (ViPR) trajectory aligns more smoothly to the blue baseline when the movement direction changes. This also indicates that the RPR component is necessary to generalize to unknown trajectories and to compensate scaling errors. 


\textbf{Dynamic motion evaluation.} In contrast to the other datasets, the \textit{Industry Scenario \#3} includes fast, large-scale, and high dynamic ego-motion in both training and test datasets. However, all estimators result in similar findings as \textit{Scenario \#2} as both scenarios embed motion dynamics and unknown trajectory shapes. Accordingly, ViPR again improves the position accuracy by \underline{\SI{17.41}{\percent}} on average (min.: \SI{11.47}{\percent}; max.: \SI{20.64}{\percent}), but this time exhibits very similar orientation errors. Improved orientation accuracy compared to \textit{Scenario \#2} is likely due to diverse orientations available in this dataset's training.


Fig.~\ref{image_com_pose_fuse_18} shows exemplary results that visualize how ViPR handles especially motion changes and motion dynamics (see the abrupt direction change between $x\in[8-9]\,m$ and $y\in[14-16]\,m$). The results also indicate that ViPR predicts the smoothest and most accurate trajectories on unknown trajectory shapes (compare the trajectory segments between $x\in[11-12]\,m$ and $y\in[14-16]\,m$). We think the reason why ViPR significantly outperforms APR by \SI{20.13}{\percent} here is because of the synergy of APR, RPR, and PE. RPR contributes most in fast motion-changes, i.e., in motion blur situation. The success of RPR may also indicate that RPR differentiates between ego- and feature-motion to more robustly estimate a pose.

\subsection{Evaluation of the RPR-Network}
\label{chapter_relnet_eval}%

\begin{figure}[!t]
    \centering
    \includegraphics[width=1\linewidth]{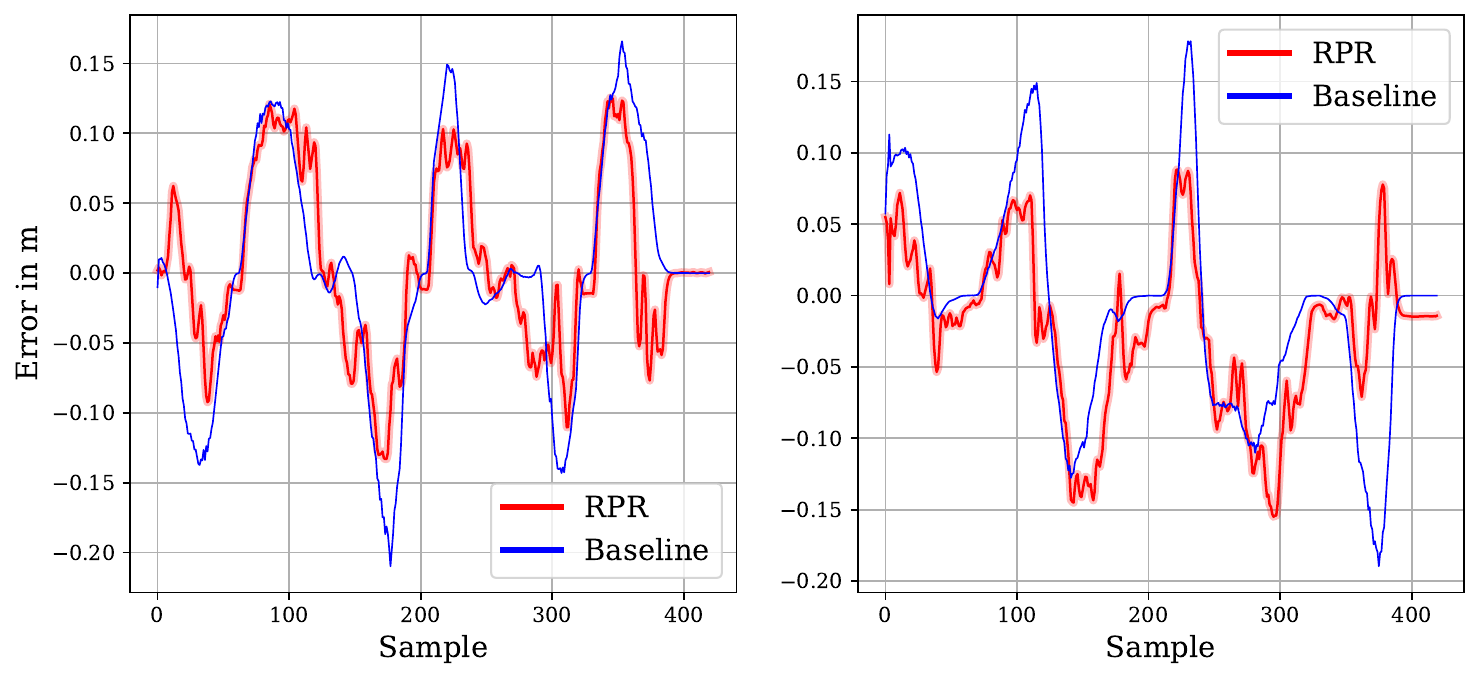}
    \captionsetup{aboveskip=-0.5pt}
    \vspace{-4mm}
    \caption{Exemplary {\textcolor{red}{RPR-results}} (displacements $m$) against the {\textcolor{blue}{baseline}} (ground truth) on the \textit{Scenario \#3} dataset (see Fig.~\ref{dataset_4_angles_test}).}
    \label{image_relative_eval}
    \vspace{-0.5cm}
\end{figure}

We use the smaller FlowNet2-s~\cite{flownet2} variant of \texttt{FlowNet2.0} as this has faster runtimes (140\,\textit{Hz}), and use it pretrained on the FlyingChairs~\cite{flownet}, ChairsSDHom and FlyingThings3D datasets. To highlight that RPR contributes to the accuracy of the final pose estimates of ViPR, we explicitly test it on the \textit{Industry Scenario \#3} that embeds dynamic motion of both ego- and feature-motion. The distance between consecutive images is up to 20\,$cm$, see Fig.~\ref{image_relative_eval}. This results in a median error of 2.49\,$cm$ in $x$- and 4.09\,$cm$ in $y$-direction on average (i.e., the error is between \SI{12.5}{\percent} and \SI{20.5}{\percent}). This shows that the RPR yields meaningful results for relative position regression in a highly dynamic and difficult setting. It furthermore appears to be relatively robust in its predictions despite both ego- and feature-motion.

\subsection{Discussion}

\textbf{6DoF Pose Regression with LSTMs.} APR-only increases the positional accuracy over \texttt{PoseNet} for all datasets, see Tab.~\ref{table_evaluation}. We found that the position errors increase when we use methods with independent and single-layer LSTM-extensions~\cite{walch_lstm,deepvo_vo,contextualnet,seifi} on both the \texttt{7-Scenes} and the \textit{Industry} datasets, by 0.04\,$m$ up to 2.07\,$m$. This motivated us to investigate stacked LSTM-layers only for the RPR- and PE-networks. We support the statement of Seifi et al.~\cite{seifi} that the motion between consecutive frames is too small, and thus, naive CNNs are already unable to embed them. Hence, additionally connected LSTMs are also unable to discover and track meaningful temporal and contextual relations between the features.

\textbf{Influence of RPR to ViPR.} To figure out the information gain of the RPR-network we also constructed ViPR in a closed end-to-end architecture through direct concatenation of the CNN-encoder-output (APR) and the LSTM-output (RPR). For a smaller OF-input ($3 \times 3$) of the RPR-model the accuracy of the \texttt{7-Scenes}~\cite{shotton_scene} dataset increases, but decreases for the \textit{Industry} dataset. This stems from the fact that the relative movements of the \texttt{7-Scenes} dataset are too small ($<2\,cm$) compared to the \textit{Industry} dataset (approx.  20\,$cm$). Hence, ViPR's contribution is limited here.

\textbf{Comparison of ViPR to state-of-the-art methods.} \texttt{VLocNet++} \cite{vlocnet2} currently achieves the best results on \texttt{7-Scenes}~\cite{shotton_scene}, but due to the small relative movement and the high ground truth error compared to \texttt{VLocNet}'s results a plausible evaluation is not possible regarding industrial applications. \texttt{MapNet}~\cite{mapnet} achieves (on average) better results than ViPR  on the \texttt{7-Scenes} dataset, but results in a similar error, e.g., 0.30\,$m$ and \SI{12.08}{\degree} on the \textit{stairs} set against ViPR's 0.31\,$m$ and \SI{12.65}{\degree}. \texttt{MapNet} has an improvement of \SI{8.7}{\percent} over ~\texttt{PoseNet2}~\cite{kendall_geometric} and achieves 41.4\,$m$ and 12.5\,$^{\circ}$ on the \texttt{RobotCar}~\cite{robotcar} dataset. However, a fair evaluation on this dataset with state-of-the-art methods requires results and code from \texttt{VLocNet}~\cite{valada_deep,vlocnet2}.

\textbf{Runtimes.}\label{chapter_runtime} The training of the APR takes 0.86\,\textit{s} per iteration for a batch size of 50 (GoogLeNet~\cite{googlenet}) on our hardware setup. The training of the RPR and PE is faster (0.065\,\textit{s}) even at a higher batch size of 100, as these models are smaller (214,605, resp.~55,239, parameters). Hence, it is possible to retrain the PE-network quickly upon environment changes. The inference time of ViPR is between 7\,\textit{ms} and 9\,\textit{ms} per sample (\texttt{PoseNet}: avg.~5\,\textit{ms}, \texttt{FlowNet2-s}: avg.~9\,\textit{ms}). In addition, ViPR does not require domain knowledge to provide scenario-dependent applicability, nor does it need a compute-intensive matcher like brute force or \texttt{RANSAC}~\cite{sweeney,brachman_dsac}. However, instead of \texttt{PoseNet}, ViPR can also use such classical approaches in its modular process pipeline. \texttt{DenseVLAD}~\cite{torii} and classical approaches are 10x (200-350\,\textit{ms}/sample) more computationally intensive than today's deep pose regression variants.

\section{Conclusion}\label{chapter_conclusion}
    
In this paper, we addressed typical challenges of learning-based visual self-localization of a monocular camera. We introduced a novel DL-architecture that makes use of three modules: an absolute and a relative pose regressor module, and a final regressor that predicts a 6DoF pose by concatenating the predictions of the two former modularities. To show that our novel architecture improves the absolute pose estimates, we compared it with a publicly available dataset and proposed novel \textit{Industry} datasets that enable a more detailed evaluation of different (dynamic) movement patterns, generalization, and scale transitions.

\section*{Acknowledgements}

\noindent
This work was supported by the Bavarian Ministry for Economic Affairs, Infrastructure, Transport and Technology through the Center for Analytics-Data-Applications (ADA-Center) within the framework of "BAYERN DIGI-TAL II".
    
{\small
\bibliographystyle{ieee_fullname}
\bibliography{egpaper_for_review}
}

\end{document}